\documentclass{article} 
\PassOptionsToPackage{table}{xcolor} 
\usepackage[utf8]{inputenc} 
\usepackage[T1]{fontenc}    
\usepackage{iclr2026_conference,times}
\usepackage{graphicx}     

\usepackage{amsmath} 
\usepackage{amssymb}
\usepackage{amsfonts}       
\usepackage{amsthm}
\usepackage{booktabs}     
\usepackage{tabularx}
\usepackage{algorithm}
\usepackage{algpseudocode}
\usepackage{algorithmicx}
\usepackage{multirow}     
\usepackage{makecell}     
\usepackage{siunitx}
\usepackage{wrapfig}
\usepackage{hyperref}       
\usepackage{url}            
\usepackage{nicefrac}       
\usepackage{microtype}      
\usepackage{pifont}  
\usepackage{xspace}
\usepackage{caption}
\usepackage{listings}
\usepackage{upquote}
\usepackage[most]{tcolorbox}

\definecolor{promptbg}{HTML}{F8F8FA}
\definecolor{promptrule}{HTML}{C3C8D0}
\definecolor{promptslot}{HTML}{A4262C}
\definecolor{promptrole}{HTML}{1F4E79}

\lstdefinestyle{promptstyle}{%
  basicstyle=\ttfamily\scriptsize,
  breaklines=true,
  breakatwhitespace=false,
  breakindent=0pt,
  postbreak=\mbox{\textcolor{promptrule}{$\hookrightarrow$}\space},
  columns=fullflexible,
  keepspaces=true,
  showstringspaces=false,
  upquote=true,
  extendedchars=false,
  literate={\{history\}}{{{\bfseries\color{promptslot}\{history\}}}}{9}
           {\{observation\}}{{{\bfseries\color{promptslot}\{observation\}}}}{13}
           {\{task\}}{{{\bfseries\color{promptslot}\{task\}}}}{6}
           {\{k\}}{{{\bfseries\color{promptslot}\{k\}}}}{3}
           {\{policy\}}{{{\bfseries\color{promptslot}\{policy\}}}}{8}
           {\{tools\}}{{{\bfseries\color{promptslot}\{tools\}}}}{7}
           {\{ending\}}{{{\bfseries\color{promptslot}\{ending\}}}}{8}
           {[USER]}{{{\bfseries\color{promptrole}[USER]}}}{6}
           {[ASSISTANT]}{{{\bfseries\color{promptrole}[ASSISTANT]}}}{11},
}

\newtcblisting{promptbox}[1]{%
  listing only,
  listing style=promptstyle,
  enhanced,
  breakable,
  colback=promptbg,
  colframe=promptrule,
  coltitle=black,
  colbacktitle=promptrule!30,
  fonttitle=\bfseries\footnotesize,
  title={#1},
  boxrule=0.4pt,
  arc=2pt,
  left=4pt,
  right=4pt,
  top=3pt,
  bottom=3pt,
}

\newsavebox{\TabA}
\newsavebox{\TabB}
\newsavebox{\TabC}
\definecolor{rowgray}{gray}{0.94}              
\definecolor{ctmrowcolor}{gray}{0.92}
\definecolor{bestgray}{gray}{0.88}

\newcommand{\xhdr}[1]{{\noindent\bfseries #1}.}
\newcommand{\name}{\textsc{CoPE}\xspace}
\definecolor{bestgray}{gray}{0.78}     
\definecolor{secondgray}{gray}{0.91}   
\newcommand{\best}{\cellcolor{bestgray}}
\newcommand{\second}{\cellcolor{secondgray}}
\DeclareRobustCommand{\hlbest}[1]{{\setlength{\fboxsep}{1.2pt}\colorbox{bestgray}{\strut #1}}}
\DeclareRobustCommand{\hlsecond}[1]{{\setlength{\fboxsep}{1.2pt}\colorbox{secondgray}{\strut #1}}}
\definecolor{groupgray}{gray}{0.95}   

\title{Co-Evolving Acting and Forecasting \\for Language Agents}

\iclrfinalcopy

\author{%
Haofei Yu$^{1}$\footnotemark[1] \quad
Yixing Chen$^{2}$\footnotemark[1] \quad
Yining Zhao$^{1}$ \quad
Xuyan Ye$^{1}$ \quad
Yuxiang Liu$^{1}$\\
\textbf{Zhigang Hua}$^{3}$ \quad
\textbf{Yan Xie}$^{3}$ \quad
\textbf{Shuang Yang}$^{3}$ \quad
\textbf{Jiaxuan You}$^{1}$ \\[4pt]
$^{1}$University of Illinois Urbana-Champaign \quad
$^{2}$University of California San Diego \\
$^{3}$Meta AI
}

\begin{document}

\maketitle
\begingroup
\renewcommand{\thefootnote}{*}
\footnotetext{Equal contribution.}
\endgroup

\begin{abstract}
Training language agents is bottlenecked by the high cost of long-horizon rollouts, yet standard agentic reinforcement learning (RL) learns only from a terminal scalar reward, leaving the supervision in intermediate steps largely unused. Prior work exploits this supervision through value targets that estimate per-step success probabilities or observation targets that fit environment response. Both are hard to construct well: (1) value targets come from the same single scalar signal for each step in the trajectory, and (2) observation targets, such as web pages and terminal logs, are long and noisy and reach only one step ahead. We propose \name, which adds a third kind of multi-step action targets. On successful on-policy trajectories, \name trains the policy backbone to generate the action sequences that follows each history. Each action is also supervised when forecast from the preceding histories, forming a temporal ensemble that the shared backbone absorbs during training, and this enables acting and forecasting to co-evolve. The policy still acts one step at a time at inference, with no overhead. Across five benchmarks (ALFWorld, ScienceWorld, WebShop, $\tau^2$-bench, and AppWorld), \name (1) achieves stronger aggregate performance than GRPO and RL baselines with predictive auxiliary targets across simulated environments, while improving overall performance on both realistic applications; (2) delivers up to $1.9\times$ and $2.0\times$ higher sample efficiency than ECHO and GRPO, respectively, on ScienceWorld; and (3) achieves these gains with a single policy backbone, without additional models, separate pretraining, or inference overhead. The codebase is available at \url{https://github.com/ulab-uiuc/CoPE}.
\end{abstract}

\section{Introduction}
\label{sec:intro}

Large language models (LLMs) enable autonomous agents to tackle complex, long-horizon tasks such as web navigation~\citep{Zheng2024GPT4VisionIA,Deng2023Mind2WebTAA,Zhang2024WebPilotAVA}, agentic coding~\citep{Yang2024SWEagentAIA,Hong2023MetaGPTMPA,Sapkota2025VibeCV}, and scientific discovery~\citep{zheng2025fromata,Buehler2024GenerativeROA,Lu2024TheAS,Yamada2025TheAS}. However, training these agents remains computationally expensive. Because each rollout involves numerous interaction steps and tool calls, extracting the maximum learning signal from a limited number of interactions is crucial~\citep{Liu2025TrulySAA}. Yet, standard agentic reinforcement learning (RL) often learns only from a single scalar reward at the end of a trajectory, leaving stepwise supervision in the process, such as intermediate actions and environment observations, largely unused.

To exploit more intermediate supervision from the process, prior work constructs two kinds of additional targets. \emph{Value targets} are estimates of how likely each intermediate step is to lead to success, obtained from process reward models~\citep{choudhury2025process,xi2026agentprm} or step-level credit assignment~\citep{feng2025gigpo,kong2026infopo}, and serve as dense rewards or advantages.
\emph{Observation targets} are the environment's next observation or state. Some methods fit a separate LLM world model to them and use it to plan or to synthesize additional experience~\citep{chen2025dreamgym,dynaweb2026}; others fit them as an auxiliary objective of the policy backbone, before or alongside policy optimization~\citep{shrivastava2026echo,paw2026,zhang2025earlyexperience,rwml2026}.


\begin{figure}[t] 
    \centering
    \includegraphics[width=\linewidth]{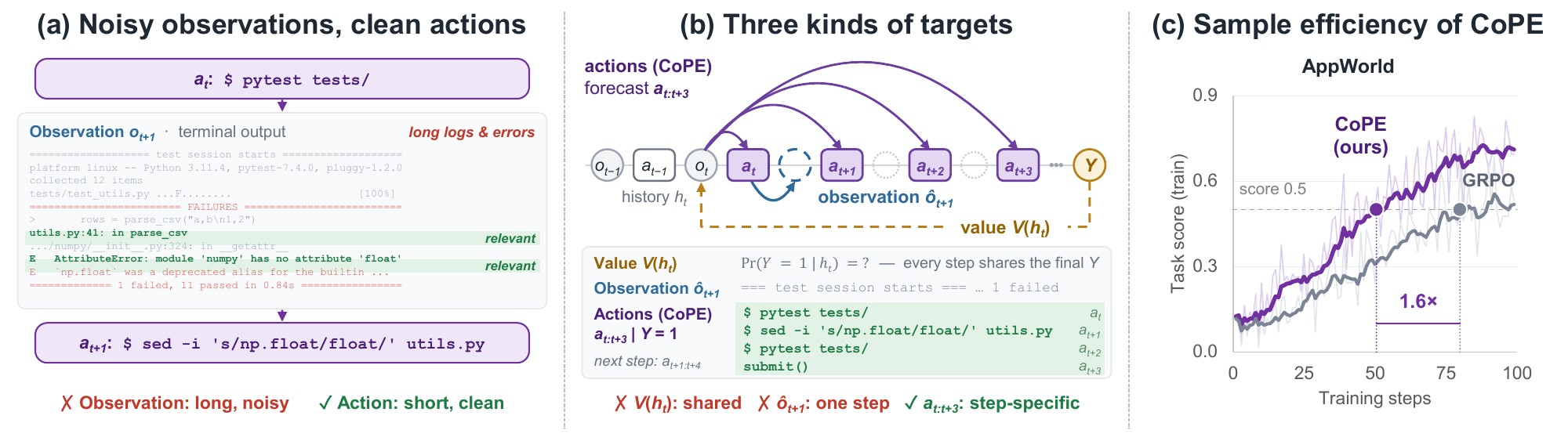}
    \vspace{-5mm}
    \caption{\textbf{Motivation of \name for sample-efficient RL.}
(\textbf{a}) A terminal observation contains many task-irrelevant tokens, whereas the corresponding action is short and decision-relevant.
(\textbf{b}) Three kinds of targets at the same history $h_t$. Value targets are shared across steps: every step of a trajectory is labeled with the same final outcome $Y$. The observation target $\hat o_{t+1}$ is noisy and reaches only one step ahead. \name's action targets condition on $Y=1$ and are specific to each step: $h_t$ is labeled with the actions that actually followed it, $a_{t:t+K}$ ($K=3$ here), and the next step with $a_{t+1:t+K+1}$.
(\textbf{c}) On AppWorld, \name reaches a training task score of 0.5 with $1.6\times$ fewer RL steps than GRPO.}
\label{fig:intro}
    \vspace{-8mm}
\end{figure}

\xhdr{Key challenges} Both existing prediction targets are difficult to construct effectively in complex agentic environments. \textbf{(1)} Value targets are derived from only the shared outcome. A trajectory records only its final outcome, not the success probability of each step, so every step is labeled with this same outcome, whose randomness comes mostly from later actions and observations rather than from the step itself. Distinguishing between steps requires step-level annotations, additional rollouts from intermediate states, or exact recurrence of states across rollouts.
\textbf{(2)} Observation targets are noisy and reach only one step ahead. In web navigation or agentic coding, most tokens of the next observation (e.g., HTML DOM trees or verbose terminal logs) are irrelevant to the decision (Fig.~\ref{fig:intro}a), so the prediction loss is dominated by reproducing them. The target also covers only single-step transition; reaching later steps requires chaining predictions, where errors compound.

\xhdr{Our solution} We propose \name, which constructs a third kind of target: \emph{multi-step action targets}. Alongside GRPO, \name employs \textbf{multi-step action forecasting} on successful on-policy trajectories, predicting actions $a_{t:t+K}$ directly from the current history while skipping the intervening observations. We show that the population optimum of this objective is $\Pr(a_{t:t+K}\mid h_t,\text{success})$, a hindsight distribution~\citep{harutyunyan2019hindsightcreditassignment,russo2026success} that conditions on both the history and the episode's eventual success. Success thus enters as a condition that selects trajectories, not as a label assigned to each step. Therefore, \name overcomes previous limitations in three key ways. \textbf{(1)} Action targets are specific to each step. A successful trajectory labels each step with its own continuation, the actions that actually followed it.  \textbf{(2)} Action targets are decision-relevant. Actions are concise (Fig.~\ref{fig:intro}a), so the loss is not spent on irrelevant observation content. \textbf{(3)} Multi-step action targets span the horizon and form an implicit temporal ensemble. Each action is supervised both where it is taken and when it is forecast from the $K$ preceding histories, and the shared backbone absorbs this ensemble into the policy through training alone. This agrees with the delay view of action chunking in robot learning, where predicting chunks trains each action from the current and several earlier observations; the resulting ensemble of delayed predictors, which ACT averages explicitly at test time~\citep{zhao2023act}, accounts for much of chunking's benefit~\citep{lazzati2026does}. \name differs by \textbf{still acting one step at a time at inference, with no overhead}.


\xhdr{Main findings} We evaluate \name on five benchmarks spanning three simulated environments (ALFWorld~\citep{shridhar2020alfworld}, ScienceWorld~\citep{wang2022scienceworld}, and WebShop~\citep{Yao2022WebShopTSA}) and two realistic applications ($\tau^2$-bench~\citep{barres2025tau2} and AppWorld~\citep{Trivedi2024AppWorldAC}). Our experiments yield three main findings. \textbf{(1)} Broad improvements: \name outperforms GRPO~\citep{shao2024deepseekmath} and observation-prediction baselines (ECHO~\citep{shrivastava2026echo} and RWML~\citep{rwml2026}) on aggregate metrics in the three simulated environments, and improves overall metrics on both realistic applications (despite a negative result on the Airline domain of $\tau^2$-bench). \textbf{(2)} Sample efficiency: on ScienceWorld, \name is up to $1.9\times$ and $2.0\times$ more sample-efficient than ECHO and GRPO, respectively. \textbf{(3)} Low overhead: on ScienceWorld, the action forecasting pass adds just $11\%$ time per training step on average, while shorter rollouts reduce the measured end-to-end step time by $9\%$; \name requires neither a separate model nor a pretraining stage.

\vspace{-2mm}
\section{Related Works}
\vspace{-2mm}

\xhdr{Sample-efficient agentic RL} Methods for sample-efficient agentic RL follow two main directions. The first makes better use of outcome rewards, either by refining credit assignment within trajectories~\citep{feng2025gigpo,Wang2025HarnessingUE,luo2025agentlightning,wang2026beacon,kong2026infopo} or by converting completed interactions into extra policy supervision via hindsight rewriting or on-policy distillation~\citep{hu2025echo,yang2026opid,wu2026seed,zheng2026grsd}. These methods change how sparse outcomes are assigned or reused, without necessarily introducing a predictive objective. The second learns from the environment feedback available at every step, typically by adding auxiliary prediction of future states or observations to policy training~\citep{zhang2025earlyexperience,shrivastava2026echo,paw2026}. Such objectives provide dense token-level supervision but generally focus on a single environment transition. \name combines elements of both directions: success filtering reuses outcome rewards, while multi-horizon prediction learns from how trajectories unfold in action space rather than reconstructing observations.

\xhdr{Multi-horizon prediction} \name is closely related to several forms of multi-horizon and outcome-conditioned prediction. Multi-token prediction trains language models to predict several future tokens~\citep{gloeckle2024mtp}, while robotic policies such as ACT and Diffusion Policy predict sequences of future actions~\citep{zhao2023act,chi2023diffusionpolicy}. Multi-token prediction is closest architecturally because multiple offsets are learned by one backbone, whereas action chunking is closest in the structure of its targets. Outcome-conditioned sequence modeling conditions policies on desired returns, while self-imitation emphasizes high-return trajectories~\citep{chen2021decisiontransformer,oh2018selfimitation}; both connect policy learning to trajectory-level outcomes. Accordingly, the $k=0$ term in \name is success-filtered imitation, while the $k\geq1$ terms extend this supervision to multiple future action offsets. \name uses these predictions as an auxiliary objective during on-policy GRPO: targets are regenerated from successful rollouts, and the predicted future actions are not executed as a chunk but affect acting through the shared backbone.
\vspace{-1mm}
\section{Preliminaries}
\vspace{-1mm}
\label{sec:prelim}

\xhdr{Language agents}
We consider a long-horizon task in which an LLM policy $\pi_\theta$ interacts with an environment for $T$ steps. At step $t$, the agent holds the history $h_t=(o_{\leq t},a_{<t})$, samples an action $a_t\sim\pi_\theta(\cdot\mid h_t)$, and receives an observation $o_{t+1}\sim P(\cdot\mid h_t,a_t)$ from the unknown environment transition $P$, so that $h_{t+1}=(h_t,a_t,o_{t+1})$. This produces a trajectory $\tau=(o_0,a_0,o_1,\ldots,a_{T-1},o_T)$. We focus on sparse-reward settings (See Appendix~\ref{app:training-rewards}), where the agent receives only a terminal success signal $Y(\tau)\in\{0,1\}$, and policy learning maximizes $J(\theta)=\mathbb{E}_{\tau\sim(\pi_\theta,P)}\big[Y(\tau)\big].$

\xhdr{Policy optimization}
A common choice for optimizing $J(\theta)$ is Group Relative Policy Optimization (GRPO)~\citep{shao2024deepseekmath}, which avoids training a critic. For each task, GRPO samples a group of $G$ trajectories $\{\tau_i\}_{i=1}^{G}$ from the rollout policy $\pi_{\mathrm{old}}$ and assigns each the normalized advantage $\hat A_i=(Y(\tau_i)-\mu)/(\sigma+\delta)$, where $\mu$ and $\sigma$ are the mean and standard deviation of rewards in the group. With the per-step ratio $\rho_{i,t}(\theta)=\pi_\theta(a_{i,t}\mid h_{i,t})/\pi_{\mathrm{old}}(a_{i,t}\mid h_{i,t})$, GRPO minimizes
\begin{equation}
\mathcal{L}_{\mathrm{GRPO}}(\theta)
=
-\frac{1}{G}
\sum_{i=1}^{G}
\frac{1}{T_i}
\sum_{t=0}^{T_i-1}
\min
\Big(
\rho_{i,t}(\theta)\hat A_i,\,
\operatorname{clip}
\big(
\rho_{i,t}(\theta),
1-\epsilon,
1+\epsilon
\big)\hat A_i
\Big).
\label{eq:grpo-loss}
\end{equation}
Because the reward evaluates trajectory as a whole, every action in $\tau_i$ receives the same advantage $\hat A_i$. The learning signal thus only says whether a trajectory succeeded, not how it unfolded. The intermediate steps, which record how each decision led to the next, provide no further supervision. This motivates extracting additional supervision from the temporal structure of the collected trajectories.

\vspace{-1mm}
\section{Extending Policy Learning Across Future Horizons}
\vspace{-1mm}
\label{sec:policy-space-wm}

Under binary outcome rewards, standard policy gradients supervise each history
$h_t$ using only the immediate next action ($K=0$), providing no direct signal
to plan further ahead. We define the \emph{hindsight action chunk}, show that
the idealized GRPO update fits only its single-step marginal ($K=0$), and
extend supervision to multi-step horizons ($K>0$) via action forecasting.

\xhdr{Hindsight action chunk}
Fix a task and a rollout policy $\pi$. Let $\Pr_\pi$ denote the rollout
distribution under $\pi$, let $Y\in\{0,1\}$ indicate whether a rollout
succeeds, and let $\Pr^{+}\triangleq\Pr_\pi(\cdot\mid Y=1)$ denote the
hindsight distribution over successful rollouts. Given a history $h_t$, the
\emph{hindsight action chunk} $q(a_{t:t+K}\mid h_t)$ characterizes which future
action sequences lead to task completion from $h_t$. Formally, it is the
distribution over the next $K+1$ actions under $\Pr^{+}$, with intermediate
observations marginalized out:
\begin{equation}
q(a_{t:t+K}\mid h_t)\;\triangleq\;\Pr\nolimits^{+}(a_{t:t+K}\mid h_t),
\qquad a_{t:t+K}=(a_t,\ldots,a_{t+K}),
\label{eq:success-chunk}
\end{equation}
which reduces to the single-step distribution $q(a_t\mid h_t)$ when $K=0$
(with actions after episode termination padded with a terminal token).

\begin{figure}[t]
\centering
\includegraphics[width=0.98\textwidth]{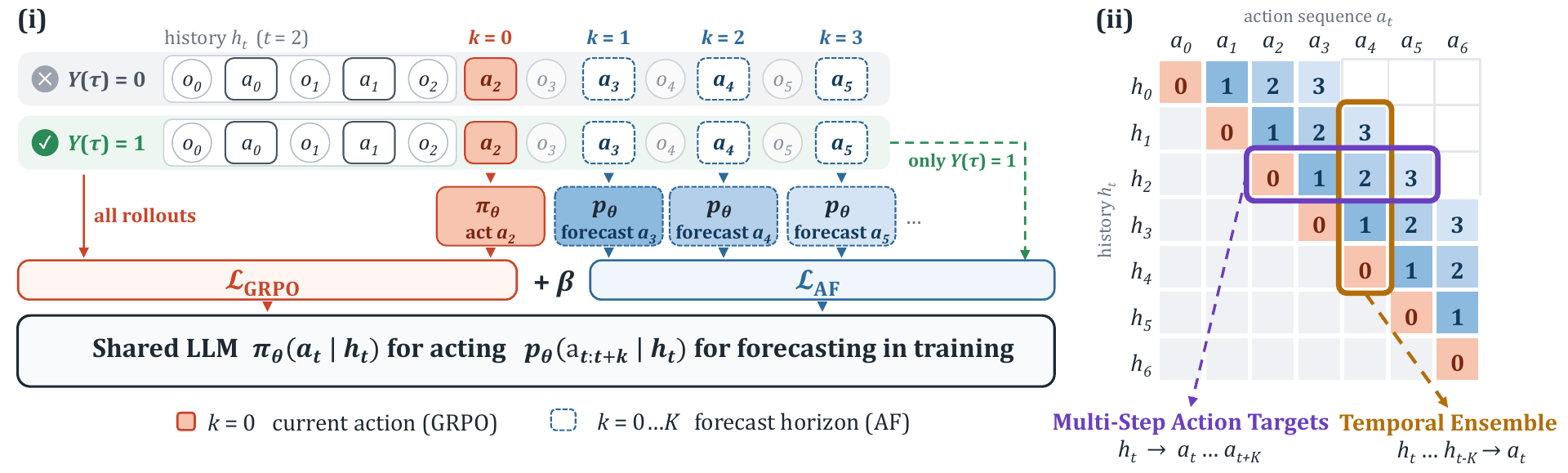}
\caption{\textbf{Overview of \name.} \textbf{(i)} Joint training of acting and forecasting. One LLM backbone is shared by both, and $\mathcal L_{\mathrm{AF}}$ fits the successful action chunk $a_{t:t+K}$. \textbf{(ii)} Co-evolution of acting and forecasting. Acting helps forecasting via trajectory collection, and forecasting helps acting via implicit temporal ensembling.}
\vspace{-10pt}
\label{fig:method}
\end{figure}

\xhdr{GRPO fits the single-step hindsight distribution}
Consider an idealized on-policy GRPO~\citep{shao2024deepseekmath} update without
KL or length normalization~\citep{liu2025understanding} (see
Alg.~\ref{alg:coevolve} for our full implementation). Since the on-policy score
has zero mean,
$\mathbb{E}_{\Pr_\pi}[\sum_t\nabla_\theta\log\pi_\theta(a_t\mid h_t)]=0$,
penalizing failed rollouts points in the same expected direction as reinforcing
successful ones. Up to a task-dependent positive scalar, the expected GRPO
update therefore follows the gradient of the single-step objective
\begin{equation}
\nabla_\theta\mathcal{L}_{\mathrm{GRPO}}
\;\propto\;
-\nabla_\theta\mathbb{E}_{\Pr^{+}}\!\Big[\textstyle\sum_t\log\pi_\theta(a_t\mid h_t)\Big]
=
\nabla_\theta\mathbb{E}_{\Pr^{+}}\!\Big[\textstyle\sum_t
\mathrm{CE}\big(q(a_t\mid h_t),\pi_\theta(a_t\mid h_t)\big)\Big],
\label{eq:grpo-posterior}
\end{equation}
where $\mathrm{CE}(q,p)\triangleq-\mathbb{E}_q[\log p]$, and the equality holds
because $a_t$ following $h_t$ in a successful rollout is a sample from
$q(a_t\mid h_t)$. While conditioning on $Y=1$ enables $q(a_t\mid h_t)$ to carry
long-horizon credit---allowing RL to learn multi-step behaviors---GRPO
supervises each $h_t$ only with its single-step marginal ($K=0$), never asking
the representation at $h_t$ to anticipate the subsequent actions $a_{t+1:t+K}$
realized for later success.

\xhdr{Action forecasting fits the multi-step hindsight distribution}
To supervise $h_t$ with the full horizon-$K$ chunk $q(a_{t:t+K}\mid h_t)$,
we introduce a parameterized autoregressive action forecaster
$p_\theta(a_{t:t+K}\mid h_t)$ and define the action forecasting objective as
the multi-step counterpart of Eq.~\ref{eq:grpo-posterior}:
\begin{equation}
\mathcal{L}_{\mathrm{AF}}(\theta)
\triangleq
-\mathbb{E}_{\Pr^{+}}\!\Big[\textstyle\sum_t\log p_\theta(a_{t:t+K}\mid h_t)\Big]
=
\mathbb{E}_{\Pr^{+}}\!\Big[\textstyle\sum_t
\mathrm{CE}\big(q(a_{t:t+K}\mid h_t),p_\theta(a_{t:t+K}\mid h_t)\big)\Big].
\label{eq:action-forecasting}
\end{equation}
Whereas Eq.~\ref{eq:grpo-posterior} fits only the single-step target
$q(a_t\mid h_t)$ ($K=0$), $\mathcal{L}_{\mathrm{AF}}$ trains $p_\theta$ to
match the joint horizon-$K$ target $q(a_{t:t+K}\mid h_t)$. Just as in the
single-step case, the second equality in Eq.~\ref{eq:action-forecasting} holds
because any action sequence $a_{t:t+K}$ observed after $h_t$ in a successful
rollout is already an exact sample from $q(a_{t:t+K}\mid h_t)$. This makes
multi-step fitting tractable: rather than learning an explicit world model to
simulate future states, we simply read target action chunks off successful
environment rollouts, where intermediate observations are naturally marginalized
under $\Pr^{+}$. Because $p_\theta$ shares parameters with the acting policy
$\pi_\theta$, minimizing $\mathcal{L}_{\mathrm{AF}}$ forces the policy's
representation at $h_t$ to encode the multi-step action plan that leads to
success.

\vspace{-1mm}
\section{\name: Co-evolving Acting and Forecasting}
\vspace{-1mm}
\label{sec:method}
\name{} trains a single LLM to both act and forecast its own future actions
(Fig.~\ref{fig:method}). The acting policy and the forecaster share a single
backbone (Sec.~\ref{sec:method-shared}), the forecaster is trained on
successful on-policy rollouts (Sec.~\ref{sec:method-policy-to-wm}), and the
forecasting objective is optimized jointly with GRPO so that acting and
forecasting co-evolve (Sec.~\ref{sec:method-coevolve}).

\subsection{Sharing Parameters for Acting and Forecasting}
\vspace{-1mm}
\label{sec:method-shared}
We instantiate both the acting policy $\pi_\theta$ and the action forecaster
$p_\theta$ on a single LLM backbone $\theta$: $\pi_\theta(a_t\mid h_t)$ is the
model's next-action distribution given history $h_t$, whereas
$p_\theta(a_{t:t+K}\mid h_t)$ is the distribution over action chunks generated
when a forecasting instruction is appended to $h_t$. Because of this
forecasting instruction, $p_\theta$ and $\pi_\theta$ are distinct
distributions: even at the first forecast position ($K=0$),
$p_\theta(a_t\mid h_t)$ does not generally equal $\pi_\theta(a_t\mid h_t)$.
Sharing the backbone offers two key benefits:
\textbf{(1)~No additional model.} Forecasting reuses the policy's parameters,
requiring no separate model to train, store, or serve.
\textbf{(2)~Shared representation.} Multi-step structure captured by the
forecasting objective is written directly into the representation from which
the policy acts, allowing acting and forecasting to co-evolve
(Sec.~\ref{sec:method-coevolve}).

\subsection{Multi-step Action Forecasting Loss}
\vspace{-1mm}
\label{sec:method-policy-to-wm}

At each iteration, we collect trajectories using the current policy
$\pi=\pi_{\theta_{\mathrm{old}}}$ and hold them fixed during optimization. As
shown in Sec.~\ref{sec:policy-space-wm}, every action chunk $a_{t:t+K}$
realized after $h_t$ in a successful rollout is a sample from the hindsight
action chunk $q(a_{t:t+K}\mid h_t)$, allowing us to train the forecaster by
maximum likelihood on these chunks without modeling environment dynamics.

\xhdr{Success-filtered multi-step action forecasting}
Since $q(a_{t:t+K}\mid h_t)$ is conditioned on success ($Y=1$), we train the
forecaster exclusively on successful trajectories. Fitting failed trajectories
as well would instead learn the unconditional rollout distribution
$\Pr_\pi(a_{t:t+K}\mid h_t)$, pushing the shared backbone toward behaviors that
GRPO suppresses. For each task $x$, let $\mathcal G_x$ be a group of $G$
rollouts sampled from $\pi$, and let
$\mathcal S_x=\{\tau\in\mathcal G_x:Y(\tau)=1\}$ be its successful subset. We
estimate the multi-step action forecasting loss (Eq.~\ref{eq:action-forecasting})
on each group via
\begin{equation}
\hat{\mathcal L}_{\mathrm{AF}}(\theta)
=
-\,
\mathbb E_{x,\,\mathcal G_x}
\Big[
\sum_{\tau\in\mathcal G_x}
\frac{Y(\tau)}{\max(1,|\mathcal S_x|)}
\sum_{t}
\log p_\theta(a_{t:t+K}\mid h_t)
\Big],
\label{eq:af-loss}
\end{equation}
where $h_t$ and $a_{t:t+K}$ are taken from $\tau$. Any chunk extending past the
end of $\tau$ is truncated after the terminal token in
Eq.~\ref{eq:success-chunk}, since subsequent padding is deterministic. At each
$h_t$, we append the forecasting instruction followed by the realized chunk,
computing each term in a single forward pass.

\xhdr{Group normalization of the forecasting loss}
Success filtering alone can destabilize training as the policy improves: as the
number of successful rollouts grows, an unnormalized sum would scale the
forecasting gradient with $|\mathcal S_x|$ and interfere with policy
optimization. The factor $1/\max(1,|\mathcal S_x|)$ in Eq.~\ref{eq:af-loss}
averages the loss across successful rollouts within each group, removing this
linear scaling. For a successful rollout $\tau$, the expected weight depends
only on its task $x$ (which $h_t$ determines). Hence, per task,
$\hat{\mathcal L}_{\mathrm{AF}}(\theta)$ equals $\mathcal L_{\mathrm{AF}}(\theta)$
(Eq.~\ref{eq:action-forecasting}) in expectation up to a positive scalar---mirroring
how the GRPO update relates to Eq.~\ref{eq:grpo-posterior}---and thus leaves
the per-history minimizer $q(a_{t:t+K}\mid h_t)$ unchanged.

\begin{algorithm}[t]
\caption{\name{} training algorithm}
\label{alg:coevolve}
\begin{algorithmic}[1]
\Require Shared backbone $\theta$, group size $G$, horizon $K$, weight $\lambda$
\For{each iteration}
    \State $\theta_{\mathrm{old}}\gets\theta$
    \State Collect a group of trajectories $\{\tau_i\}_{i=1}^{G}$ with $\pi_{\theta_{\mathrm{old}}}$
    \State $\hat A_i \gets (Y(\tau_i)-\mu)/(\sigma+\delta)$
    \State $\rho_{i,t}\gets \pi_\theta(a_{i,t}\mid h_{i,t})/\pi_{\theta_{\mathrm{old}}}(a_{i,t}\mid h_{i,t})$
    \State $\hat{\mathcal L}_{\mathrm{GRPO}}(\theta) \gets -\frac{1}{G}\sum_{i=1}^{G}\frac{1}{T_i}\sum_{t}\min\!\big(\rho_{i,t}\hat A_i,\ \operatorname{clip}(\rho_{i,t},1\!-\!\epsilon,1\!+\!\epsilon)\hat A_i\big)$ \Comment{Eq.~\ref{eq:grpo-loss}}
    \State $\mathcal S\gets\{i:Y(\tau_i)=1\}$
    \State $\hat{\mathcal L}_{\mathrm{AF}}(\theta) \gets -\frac{1}{\max(1,|\mathcal S|)}\sum_{i\in\mathcal S}\sum_{t}\log p_\theta(a_{i,t:t+K}\mid h_{i,t})$ \Comment{Eq.~\ref{eq:af-loss}}
    \State Update $\theta$ by minimizing $\hat{\mathcal L}_{\mathrm{GRPO}}(\theta)+\lambda\,\hat{\mathcal L}_{\mathrm{AF}}(\theta)$
\EndFor
\end{algorithmic}
\end{algorithm}

\subsection{Co-evolution of Acting and Forecasting}
\vspace{-1mm}
\label{sec:method-coevolve}
\xhdr{Acting helps forecasting via trajectory collection}
For a fixed rollout policy $\pi$ and a sufficiently expressive model,
$\mathcal L_{\mathrm{AF}}(\theta)$ is minimized on every history reached by
successful rollouts by the hindsight action chunk itself
(Eq.~\ref{eq:action-forecasting}), which by Bayes' rule is
\begin{equation}
p_{\theta^\star}(a_{t:t+K}\mid h_t)
=
q(a_{t:t+K}\mid h_t)
\;\propto\;
\Pr\nolimits_\pi(a_{t:t+K}\mid h_t)\,
\Pr\nolimits_\pi(Y=1\mid h_t,\,a_{t:t+K}).
\label{eq:af-optimum}
\end{equation}
The first factor represents the action chunks proposed by the policy $\pi$, and
the second is their success probability (value), which success filtering
enforces via rejection sampling. This reweighting keeps the target ahead of the
generating policy: acting at every history according to the single-step
marginal $q(a_t\mid h_t)$ succeeds at least as often as acting with
$\pi$~\citep{russo2026success}. Since both factors depend on the current policy
$\pi$, forecasting relies on acting to supply successful rollouts and receives
more of them as the policy improves.

\xhdr{Forecasting helps acting via implicit temporal ensembling}
Sec.~\ref{sec:policy-space-wm} views the forecasting loss along each row of
Fig.~\ref{fig:method}(ii): each history $h_t$ must anticipate its future chunk
$a_{t:t+K}$. Viewed along each column, every action $a_t$ of a successful
rollout also lies within the chunks of the $K$ preceding histories, receiving
supervision from each of $h_t,h_{t-1},\dots,h_{t-K}$ through the forecast
$p_\theta(a_t\mid h_{t-k},a_{t-k:t-1})$, rather than only from $h_t$ at the
moment it is executed as in GRPO. These forecasts do not conflict with acting:
since all are conditionals of the same hindsight distribution $\Pr^{+}$, the
optimal forecast of $a_t$ issued at $h_{t-k}$ equals the GRPO target
$q(a_t\mid h_t)$ marginalized over the intermediate observations between $t-k$
and $t$. Because the forecaster shares $\theta$ with the acting policy
(Sec.~\ref{sec:method-shared}), these earlier forecasts are implicitly
ensembled into $\pi_\theta$ through the shared representation (see
Sec.~\ref{sec:disc-co-evolution}).

\xhdr{Joint optimization and inference}
We optimize the shared parameters with
$\mathcal L(\theta)=\hat{\mathcal L}_{\mathrm{GRPO}}(\theta)+\lambda\,\hat{\mathcal L}_{\mathrm{AF}}(\theta)$,
where $\hat{\mathcal L}_{\mathrm{AF}}$ is given by Eq.~\ref{eq:af-loss} and
$\lambda$ is a weighting hyperparameter (Alg.~\ref{alg:coevolve}). \textbf{The
forecasting objective is used only during training:} at inference, the agent
acts with $\pi_\theta$ alone, generating one action per step without the
forecasting instruction, so \name{} adds no inference cost over a GRPO-trained
policy.

\vspace{-2mm}
\section{Experimental Settings}
\vspace{-2mm}
\label{sec:exp-settings}

\xhdr{Tasks} We evaluate on five long-horizon agent benchmarks with sparse rewards that differ in observation complexity, action space, and horizon. Three are simulated environments: \textbf{(1)} ALFWorld~\citep{shridhar2020alfworld} for household tasks, \textbf{(2)} ScienceWorld~\citep{wang2022scienceworld} for elementary science experiments, and \textbf{(3)} WebShop~\citep{Yao2022WebShopTSA} for online shopping over long web-page observations. Two are realistic applications: \textbf{(1)} $\tau^2$-bench~\citep{barres2025tau2} for policy-constrained customer service with a user simulator, and \textbf{(2)} AppWorld~\citep{Trivedi2024AppWorldAC} for composing API calls across consumer apps with code as actions. We use Qwen2.5-7B-Instruct~\citep{Yang2024Qwen25TR} for the three simulated environments and $\tau^2$-bench, and Qwen2.5-14B-Instruct for the more difficult AppWorld tasks. Details are available in Appendix~\ref{appendix:artifact_details}. Unless otherwise stated, all analysis experiments are carried out on ScienceWorld.

\xhdr{Baselines} On the simulated environments (Table~\ref{tab:alfworld_webshop}), we compare against three groups.
(\textbf{1}) \emph{Prompting}: Qwen2.5-7B-Instruct, Gemini-2.5-Flash-Lite~\citep{comanici2025gemini}, GPT-5-mini and GPT-5.5~\citep{OpenAI2025GPT5,OpenAI2026GPT55}, and Reflexion~\citep{Shinn2023ReflexionLA}, which adds verbal self-reflection on failed trials.
(\textbf{2}) \emph{RL w/o auxiliary loss}: RFT~\citep{yuan2023scaling}, rejection sampling fine-tuning; GRPO~\citep{shao2024deepseekmath}, critic-free RL with group-normalized advantages; and EMPG~\citep{Wang2025HarnessingUE}, which rescales GRPO's step-level gradients by policy entropy.
(\textbf{3}) \emph{RL w/ auxiliary loss}: RWML~\citep{rwml2026}, which first trains next-state prediction on environment transitions and then runs GRPO; and ECHO~\citep{shrivastava2026echo}, which adds a next-observation prediction loss to GRPO.
Both supervise only single-step transitions, testing whether forecasting actions over multiple steps is more effective than predicting the next observation.
On $\tau^2$-bench and AppWorld (Tables~\ref{tab:tau2} and~\ref{tab:appworld}), we include published baseline numbers reported by G2PO~\citep{wang2026group} and InfoPO~\citep{kong2026infopo}, together with GRPO and \name runs from our implementation. Because the published AppWorld results use a different task subset and protocol (Appendix~\ref{app:baseline_methods}), our direct comparison is between GRPO (ours) and \name under our shared setup.

\vspace{-2mm}
\section{Experimental Results}
\vspace{-2mm}
\label{sec:exp}

\textbf{\name achieves aggregate gains across five agentic benchmarks.}
Across both simulated and realistic agentic tasks, \name improves the benchmark-level aggregate metrics over GRPO. On ALFWorld (Table~\ref{tab:alfworld_webshop}), \name improves performance by 5.0 points over GRPO. The gains are larger on WebShop and ScienceWorld, where success rates increase by 12.5 points to $77.0\%$ and by 8.0 points to $72.5\%$, respectively; on ScienceWorld, \name also achieves the best performance among the RL-based methods. The same trend largely extends to more realistic benchmarks. On $\tau^2$-bench (Table~\ref{tab:tau2}), \name improves overall success from $27.5\%$ to $28.7\%$, driven by an approximately 12-point gain on Retail, despite a drop on Airline. We analyze the drop in Sec.~\ref{sec:disc-success}. Finally, on AppWorld (Table~\ref{tab:appworld}), \name substantially improves over GRPO on both splits, increasing success from $41.67\%$ to $55.36\%$ on the normal split and from $18.71\%$ to $22.54\%$ on the challenge split.

\begin{wraptable}{r}{0.42\textwidth}
\centering
\small
\vspace{-4mm}
\caption{\textbf{Results on $\tau^2$-bench}. All results are Avg@4 for task success rate with Qwen2.5-7B-Instruct.}
\label{tab:tau2}
\vspace{-2mm}

\setlength{\tabcolsep}{2pt}
\begin{tabular}{
    l
    S[table-format=2.1]
    S[table-format=2.1]
    S[table-format=2.1]
    S[table-format=2.1]
}
\toprule
\textbf{Method}
& {\textbf{Telec.}}
& {\textbf{Retail}}
& {\textbf{Air}}
& {\textbf{Overall}} \\
\midrule
\rowcolor{groupgray}\multicolumn{5}{l}{\textit{Agentic Prompting}} \\
Qwen2.5     & 14.4 & 13.1 &  7.5 & 11.7 \\
ReAct       & 13.1 & 13.8 & 10.0 & 12.3 \\
\midrule
\rowcolor{groupgray}\multicolumn{5}{l}{\textit{Published RL results}} \\
RAGEN       & 17.5 & 17.5 & 15.0 & 16.7 \\
Search-R1   & 15.6 & 15.0 & 10.0 & 13.5 \\
UserRL      & 13.8 & 16.9 &  7.5 & 12.7 \\
InfoPO      & 18.1 & \second 18.8 & \second 16.3 & 17.7 \\
\midrule
\rowcolor{groupgray}\multicolumn{5}{l}{\textit{Our controlled runs}} \\
GRPO (ours) & \second 38.8 & 17.5 & \best 26.2 & \second 27.5 \\
CoPE        & \best 40.6 & \best 29.4 & 16.2 & \best 28.7 \\
\bottomrule
\end{tabular}

\vspace{2mm}

\caption{\textbf{Results on AppWorld}. All results are success rates (\%) with Qwen2.5-14B-Instruct. $\dagger$ indicates results in G2PO.}
\label{tab:appworld}
\vspace{-2mm}

\setlength{\tabcolsep}{8pt}
\begin{tabular}{
    l
    S[table-format=2.2]
    S[table-format=2.2]
}
\toprule
\textbf{Method}
& {\textbf{Normal}}
& {\textbf{Challenge}} \\
\midrule
\rowcolor{groupgray}\multicolumn{3}{l}{\textit{Agentic Prompting}} \\
Qwen2.5 &  1.19 & 0.72 \\
ReAct & 10.71 & 2.64 \\
\midrule
\rowcolor{groupgray}\multicolumn{3}{l}{\textit{Published RL results}} \\
PPO$^\dagger$     & 19.10 & {--} \\
RLOO$^\dagger$    & 24.80 & {--} \\
GiGPO$^\dagger$   & 25.70 & {--} \\
GRPO$^\dagger$    & 24.80 & {--} \\
\midrule
\rowcolor{groupgray}\multicolumn{3}{l}{\textit{Our controlled runs}} \\
GRPO (ours) & \second 41.67 & \second 18.71 \\
CoPE        & \best 55.36 & \best 22.54 \\
\bottomrule
\end{tabular}

\vspace{-16mm}
\end{wraptable}

\textbf{\name outperforms observation-based auxiliary losses on aggregate simulated-task metrics.}
On the three simulated environments where ECHO and RWML are evaluated, \name achieves the best aggregate results. It exceeds the stronger baseline by 1.0 point on ALFWorld, by 4.4 average-score points and 5.5 success-rate points on WebShop, and by 5.0 success-rate points on ScienceWorld. These gains are not uniform across ALFWorld subtasks: ECHO or RWML performs better on Cool and Pick2.

\textbf{\name improves sample efficiency.} Figure~\ref{fig:intro} compares \name with GRPO on AppWorld, while Fig.~\ref{fig:task-source-curve} compares \name with ECHO and GRPO on ScienceWorld. Overall, \name is $1.6\times$ and $2.0\times$ more sample-efficient than GRPO on AppWorld and ScienceWorld, respectively. On ScienceWorld, ECHO requires 53 and 114 steps to reach equivalent performance milestones (making \name $1.7\times$ to $1.9\times$ faster). This acceleration is even more pronounced at higher reward tiers: \name achieves a 0.6 reward by step 86, a level neither baseline reaches within the first 150 steps. Thus, the multi-step action forecasting loss not only accelerates early-stage learning but also enables the policy to reach a higher performance level under the same budget.

\textbf{\name introduces little training overhead.}
Because \name adds a supervised pass over successful rollouts, its direct cost scales with the batch success rate. On ScienceWorld, this pass adds $11\%$ time per training step on average, while shorter rollouts reduce the measured end-to-end step time by $9\%$ compared with GRPO (Fig.~\ref{fig:training-overhead}).

\begin{table*}[t]
\centering
\small
\caption{\textbf{Evaluation results on simulated agent tasks.}
For ALFWorld, we report the success rate (\%) on each subtask and overall.
For WebShop and ScienceWorld, we report the average score and the success rate (\%).
All RL methods use Qwen2.5-7B-Instruct.
\smash{\hlbest{Dark gray}} and \smash{\hlsecond{light gray}} mark the best and second-best results in each column across all methods.
More details are available in Appendix~\ref{appendix:experimental_details}.}
\label{tab:alfworld_webshop}
\label{main-eval}
\vspace{-2mm}
\sisetup{
  round-mode          = places,
  round-precision     = 1,
  table-number-alignment = center,
}
\setlength{\tabcolsep}{2pt}
\renewcommand{\arraystretch}{1}
\begin{tabular}{
  l
  *{6}{S[table-format=3.1]}
  @{\hspace{11pt}} S[table-format=3.1]
  @{\hspace{14pt}} S[table-format=2.1] S[table-format=2.1]
  @{\hspace{14pt}} S[table-format=-2.1] S[table-format=2.1]
}
\toprule
\multirow{2}{*}[-0.4ex]{\textbf{Method}}
  & \multicolumn{7}{c}{\textbf{ALFWorld}}
  & \multicolumn{2}{c}{\textbf{WebShop}}
  & \multicolumn{2}{c}{\textbf{ScienceWorld}} \\
\cmidrule(lr){2-8} \cmidrule(lr){9-10} \cmidrule(l){11-12}
  & {Pick} & {Look} & {Clean} & {Heat} & {Cool} & {Pick2} & {\textbf{All}}
  & {Score} & {Succ.} & {Score} & {Succ.} \\
\midrule
\rowcolor{ctmrowcolor}\multicolumn{12}{l}{\textit{Agentic prompting}} \\
Gemini-2.5-flash-lite  & 82.61 & 36.84 & 37.84 & 20.00 & 21.43 & 48.89 & 46.00 & 7.83 & 6.50 & 13.85 & 15.00 \\
GPT-5-mini             & 82.61 & 47.37 & 27.03 & 36.00 & 14.29 & 75.56 & 52.00 & 11.74 & 7.00 & 40.13 & 43.50 \\
GPT-5.5                & \second 97.83 & \best 100.00 & 91.89 & 88.00 & \second 89.29 & \second 93.33 & 93.50 & 66.10 & 56.50 & \best 66.77 & \best 79.00 \\
Qwen2.5-7B             & 69.57 & 31.58 & 29.73 & 36.00 & 17.86 & 11.11 & 34.00 & 25.28 & 2.50 & -13.34 & 2.50 \\
Qwen2.5-7B + Reflexion & 84.78 & 26.32 & 51.35 & 44.00 & 32.14 & 55.56 & 54.00 & 21.93 & 4.00 & -3.69 & 6.00 \\
\midrule
\rowcolor{ctmrowcolor}\multicolumn{12}{l}{\textit{RL methods w/o auxiliary loss}} \\
RFT                    & 91.30 & 73.68 & 72.97 & 84.00 & 64.29 & 71.11 & 77.00 & 51.40 & 16.00 & -11.00 & 4.50 \\
GRPO                   & \best 100.00 & 73.68 & \second 94.59 & 84.00 & \second 89.29 & 88.89 & 90.50 & 84.07 & 64.50 & 52.78 & 64.50 \\
GRPO + EMPG            & \second 97.78 & 57.89 & 89.19 & 88.00 & \second 89.29 & \best 95.56 & 89.50 & 84.33 & 67.50 & 48.17 & 46.50 \\
\midrule
\rowcolor{ctmrowcolor}\multicolumn{12}{l}{\textit{RL methods w/ auxiliary loss}} \\
RWML + GRPO            & \best 100.00 & \second 89.47 & 89.19 & \second 92.00 & \best 96.43 & 84.44 & 92.00 & 81.16 & 64.00 & 39.35 & 65.00 \\
ECHO                   & \second 97.83 & \second 89.47 & 91.89 & \best 96.00 & \best 96.43 & \second 93.33 & \second 94.50 & \second 85.00 & \second 71.5 & 44.77 & 67.50 \\
\name                  & \best 100.00 & \best 100.00 & \best 97.30 & \best 96.00 & \second 89.29 & 91.11 & \best 95.50 & \best 89.40 & \best 77.00 & \second 53.40 & \second 72.50 \\
\bottomrule
\end{tabular}
\vspace{-3mm}
\end{table*}


\vspace{-2mm}
\section{Discussions}
\vspace{-2mm}

\textbf{Q1: What is the empirical evidence for acting and forecasting co-evolution?}
\label{sec:disc-co-evolution}

In Sec.~\ref{sec:method-coevolve}, we analyze the co-evolution mechanism between acting and forecasting. The acting-to-forecasting direction is straightforward: the policy collects successful trajectories as the forecast targets. Conversely, for the forecasting-to-acting direction, our analysis suggests that forecasting improves action selection through implicit ensembling via shared parameters. Here, we provide empirical evidence to support this claim.

\xhdr{Evidence for implicit temporal ensembling} For this experiment, we train a separate LoRA forecaster on ScienceWorld trajectories collected by the model, while the model's parameters only receive policy gradient. The policy therefore supplies the forecasting data, but forecasting gradients cannot update the policy. We then use the separately parameterized forecaster to rerank candidate actions using forecasts from multiple horizons while \name model itself acts as forecaster:
\begin{equation}
\log \pi(a\mid h_t)=(1-\eta)\log\pi_\theta(a\mid h_t)+\frac{\eta}{K}\sum_{k=1}^{K}\log p_{\theta'}(a\mid h_{t-k},\,a_{t-k:t-1}),
\end{equation}
where $K=2$ in this experiment and $\theta'$ denotes the LoRA-trained forecaster. We rank eight candidate actions by $\log\pi(a|h_t)$ and execute the highest-ranked one. At every checkpoint in Table~\ref{tab:af-lora}, explicit forecasting improves GRPO and narrows its gap to \name{}, but adds nothing to \name{}, as predicted in Sec.~\ref{sec:method-coevolve}. The forecasts thus carry information for acting that policy gradient alone does not acquire, and \name{}'s policy already contains it: what GRPO obtains only through a separate forecaster and extra forward passes at test time, \name{} delivers in a single forward pass. Forecasting therefore changes what the policy itself knows, completing the forecasting-to-acting half of the co-evolution.

\textbf{Q2: Why do multi-step targets outperform imitating successful actions?}
\label{sec:disc-k0}

\citet{lazzati2026does} traces the benefits of action chunking in robot imitation learning~\citep{zhao2023act} to delayed prediction: a prediction of $a_t$ has a delay $k$ if made without the last $k$ observations. Since a chunking policy predicts $a_t$ from $o_t, o_{t-1}, \dots$, it acts as an ensemble of delayed predictors: a single delay recovers much of the benefit but falls short on harder tasks, while an ensemble over randomized delays matches chunking.

In our setting, the $k$-th term of the forecasting loss in Eq.~\ref{eq:action-forecasting} is a delay-$k$ prediction. Each action is thus trained at every delay up to $K$, as in chunking. The variants in Table~\ref{tab:sciworld-k0-comparison} differ in which delays they train: success-filtered SFT and $K=0$ train only delay $0$, on $\pi_\theta$ directly and under the forecasting instruction (Sec.~\ref{sec:method-shared}) respectively; the invalid-action filter replaces each rejected $K=0$ target with the next valid action, a single delay set by an environment-specific signal; and \name{} trains all delays up to $K$. Relative to GRPO, they change the success rate by $-7.5$, $+2.0$, $+5.0$, and $+8.0$ points.

\begin{figure*}[t]
\centering
\captionsetup{font=footnotesize,labelfont=normalfont,textfont=normalfont,
  format=plain,justification=justified,singlelinecheck=false,
  margin=0pt,indention=0pt,skip=0pt,position=bottom}

\noindent
\begin{minipage}[t]{0.32\linewidth}
\vspace{0pt}
\includegraphics[width=\linewidth]{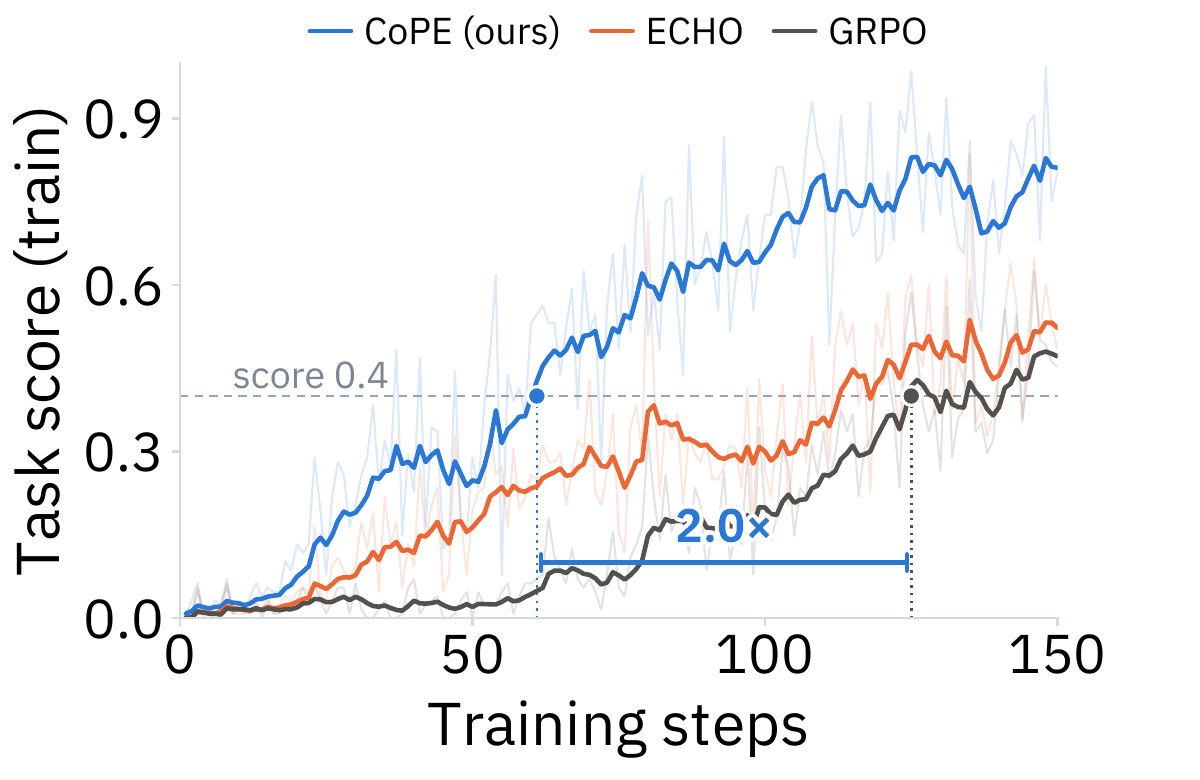}
\end{minipage}\hfill
\begin{minipage}[t]{0.32\linewidth}
\vspace{0pt}
\includegraphics[width=\linewidth]{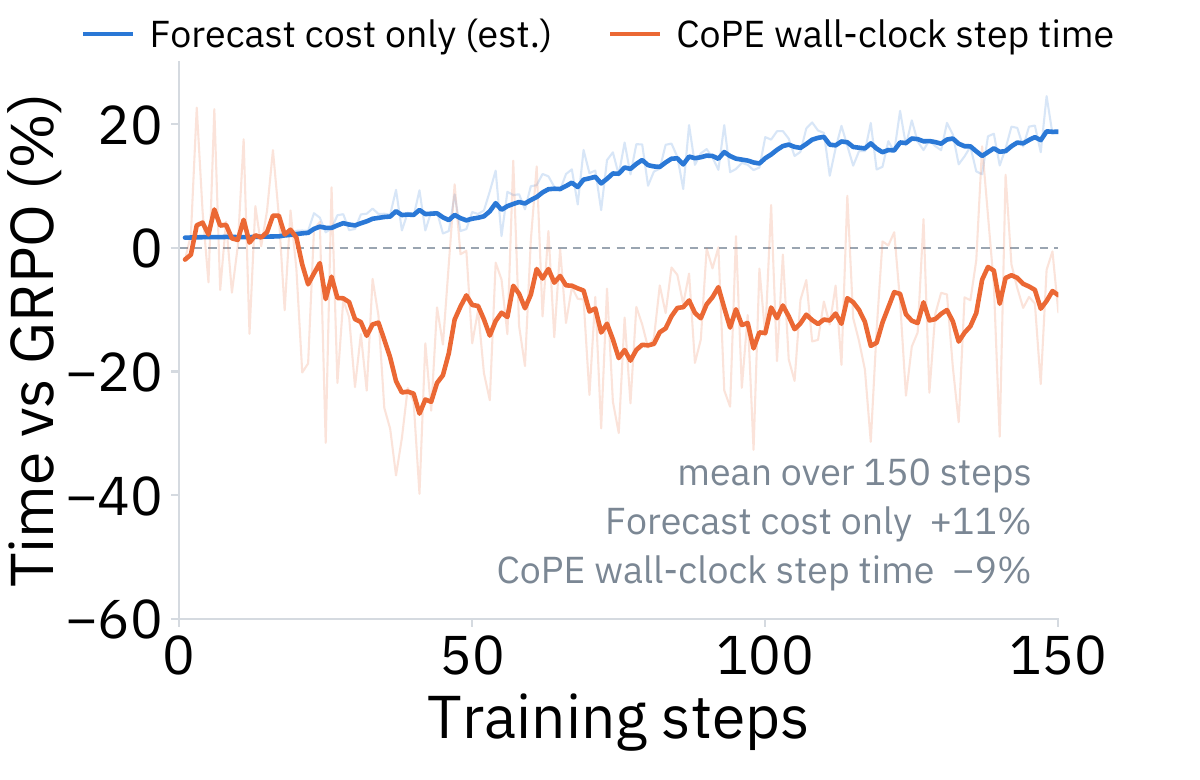}
\end{minipage}\hfill
\begin{minipage}[t]{0.32\linewidth}
\vspace{0pt}
\includegraphics[width=\linewidth]{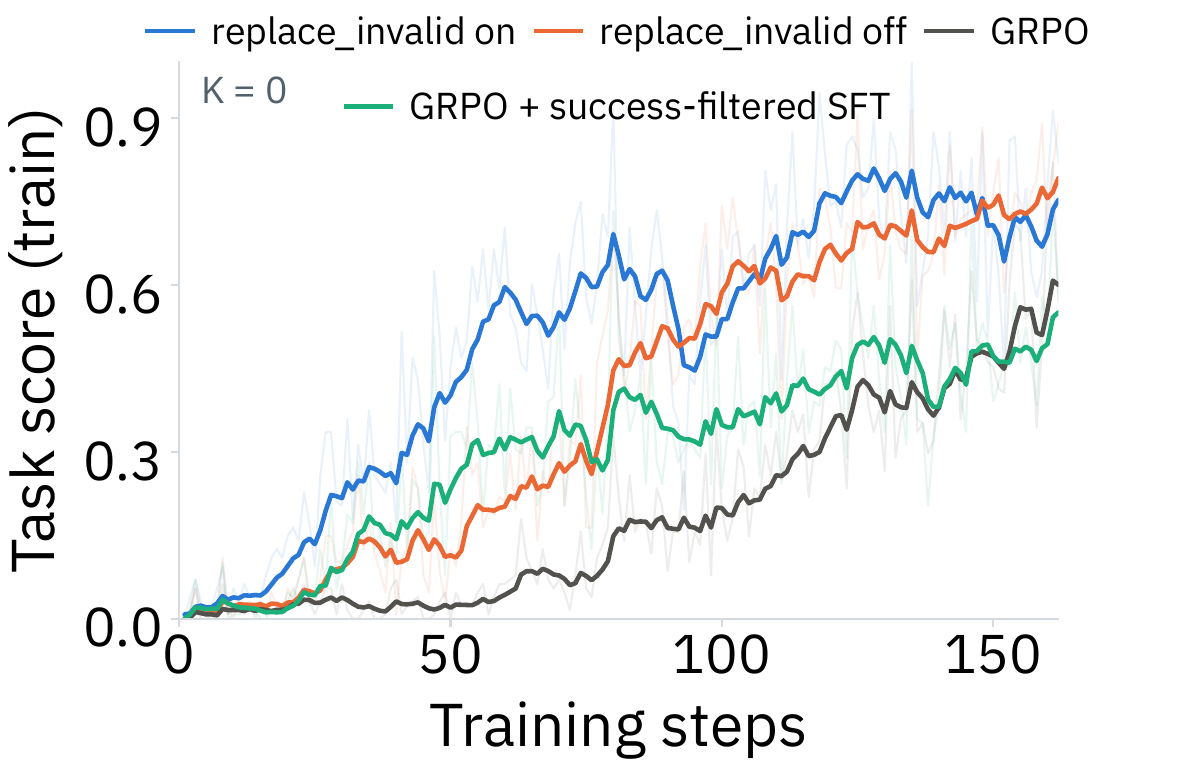}
\end{minipage}
\par\nointerlineskip\vspace{4pt}

\noindent
\begin{minipage}[t]{0.32\linewidth}
\vspace{0pt}
\caption{\textbf{Analysis of sample efficiency.}
\name reaches $0.4$ in about half the RL steps required by GRPO on ScienceWorld.}
\label{fig:task-source-curve}
\end{minipage}\hfill
\begin{minipage}[t]{0.32\linewidth}
\vspace{0pt}
\caption{\textbf{Analysis on wall-clock efficiency.}
Forecasting adds $11\%$ per-step cost, while end-to-end step time decreases by $9\%$.}
\label{fig:training-overhead}
\end{minipage}\hfill
\begin{minipage}[t]{0.32\linewidth}
\vspace{0pt}
\caption{\textbf{Invalid-action filtering.}
Replacing invalid action with the next valid one improves the sample efficiency of the
$K=0$ baseline.}
\label{fig:skip-invalid}
\end{minipage}
\par\vspace{-2mm}
\end{figure*}

\xhdr{Delay $0$ carries little information about success}
Success filtering keeps detours, suboptimal or invalid actions that the agent later recovers from, since they barely affect success. At such a step, the delay-$0$ target is the detour itself, while the recovery that decides success appears only at later delays. Imposed on $\pi_\theta$ directly, the delay-$0$ target speeds up early training but later hurts it (Fig.~\ref{fig:skip-invalid}).

\xhdr{A single delay versus an ensemble over delays}
The filter confirms that a delayed target helps ($+2.0\to+5.0$, Fig.~\ref{fig:skip-invalid}), but it needs an environment-specific signal and covers only detours that the environment rejects. \name{} needs no such signal and trains every action at all delays, reaching $+8.0$: delay $0$, a single delay, and the full ensemble rank in the same order as for robot policies.

\begin{table*}[t]
\centering
\captionsetup{font=footnotesize,labelfont=normalfont,textfont=normalfont,
  format=plain,justification=justified,singlelinecheck=false,
  margin=0pt,indention=0pt,skip=0pt,position=top}

\noindent
\begin{minipage}[t]{0.32\linewidth}
\vspace{0pt}
\caption{\textbf{Backbone ablations.}
WS/SW: WebShop/ScienceWorld.
Both backbones share training settings.
Entries: \mbox{GRPO $\to$ \name{}}.}
\label{tab:qwen3-backbone}
\end{minipage}\hfill
\begin{minipage}[t]{0.32\linewidth}
\vspace{0pt}
\caption{\textbf{Comparison of action forecasting variants.} We report success rate and gain with GRPO on ScienceWorld.}
\label{tab:sciworld-k0-comparison}
\end{minipage}\hfill
\begin{minipage}[t]{0.32\linewidth}
\vspace{0pt}
\caption{\textbf{Explicit ensembling results.} G$_{150}$ and G$_{250}$ denote checkpoints after 150 and 250 training steps on ScienceWorld.}
\label{tab:af-lora}
\end{minipage}
\par\nointerlineskip\vspace{4pt}

\scriptsize
\setlength{\tabcolsep}{2pt}
\renewcommand{\arraystretch}{1.10}
\noindent
\begin{minipage}[t]{0.32\linewidth}
\vspace{0pt}
\begin{tabular*}{\linewidth}{@{\extracolsep{\fill}}ll r@{\,$\to$\,}l r@{\,$\to$\,}l@{}}
\toprule
\multicolumn{2}{@{}l}{Benchmark} & \multicolumn{2}{c}{Qwen2.5-7B} & \multicolumn{2}{c@{}}{Qwen3-8B} \\
\midrule
\multirow{2}{*}{WS} & Score & 84.1 & \textbf{89.4} & 83.4 & \textbf{87.4} \\
 & Succ. & 64.5 & \textbf{77.0} & 64.0 & \textbf{72.5} \\
\multirow{2}{*}{SW} & Score & 52.8 & \textbf{53.4} & 50.1 & \textbf{58.3} \\
 & Succ. & 64.5 & \textbf{72.5} & 61.5 & \textbf{71.5} \\
\bottomrule
\end{tabular*}
\end{minipage}\hfill
\begin{minipage}[t]{0.32\linewidth}
\vspace{0pt}
\begin{tabular*}{\linewidth}{@{\extracolsep{\fill}}lcc@{}}
\toprule
Method & Succ. & $\Delta$ \\
\midrule
GRPO + success-filtered SFT & 57.0 & $-$7.5 \\
$K{=}0$ (no invalid-action filter) & 66.5 & +2.0 \\
$K{=}0$ (invalid-action filter) & 69.5 & +5.0 \\
\name{} ($K{=}2$) & \textbf{72.5} & \textbf{+8.0} \\
\bottomrule
\end{tabular*}
\end{minipage}\hfill
\begin{minipage}[t]{0.32\linewidth}
\vspace{0pt}
\begin{tabular*}{\linewidth}{@{\extracolsep{\fill}}lccc@{}}
\toprule
Setting & $G_{150}$ & $G_{250}$ & \name{} \\
\midrule
Top-1 sampling & 56.5 & 64.5 & 72.5 \\
Top-8 ($\eta{=}0$) & 62.5 & 69.0 & \textbf{73.5} \\
Top-8 ($\eta{=}0.75$) & 70.0 & 72.0 & 73.0 \\
$\Delta$ ($\eta{=}0.75$ vs.\ $0$) & +7.5 & +3.0 & $-$0.5 \\
\bottomrule
\end{tabular*}
\end{minipage}
\par
\end{table*}

\textbf{Q3: Why must action forecasting loss be conditioned on success, and what does it learn?}
\label{sec:disc-success}

Fundamentally, the action forecasting loss must align with the primary RL objective. Without success conditioning, the model is forced to imitate average or failed trajectories, creating a severe objective conflict and leading to zero reward after the first few steps of training. With it, the forecaster learns how the agent behaves in episodes that eventually succeed. Because success depends on the task as well as on the agent's choices, these episodes may over-represent easy tasks, a known limitation of outcome-conditioned imitation~\citep{paster2022you,eysenbach2022imitating}.

\xhdr{Mixed results on $\tau^2$-bench}
\name{} gains on Retail ($17.5\%\to29.4\%$) but drops on Airline ($26.2\%\to16.2\%$), because Airline's task and reward design makes this over-representation extreme. The reward checks state-based assertions on the final database, and 38\% of Airline tasks expect the database to stay unchanged, against 4.4\% in Retail. On these tasks, refusing the request is fully rewarded, and they are by far the easiest since the policy does not need to do anything: early in training they succeed 52\% of the time, against 8.9\% for tasks that require changing the database, so most successful episodes, the only ones the forecasting loss learns from, are refusals. Yet a refusal task and an action task differ only in details: a request to cancel a booking must instead be refused once the customer demands a refund above 50\%. A rewarded refusal and an unrewarded one thus share nearly the same history, and the forecaster, trained only on the rewarded copies, carries refusals over to tasks that require action. These tasks then succeed even less often, which skews the targets further toward refusal.

\begin{figure*}[t]
    \centering
    \begin{minipage}[t]{0.32\textwidth}
        \centering
        \vspace{0pt}
            \makebox[\linewidth][c]{\includegraphics[width=\linewidth]{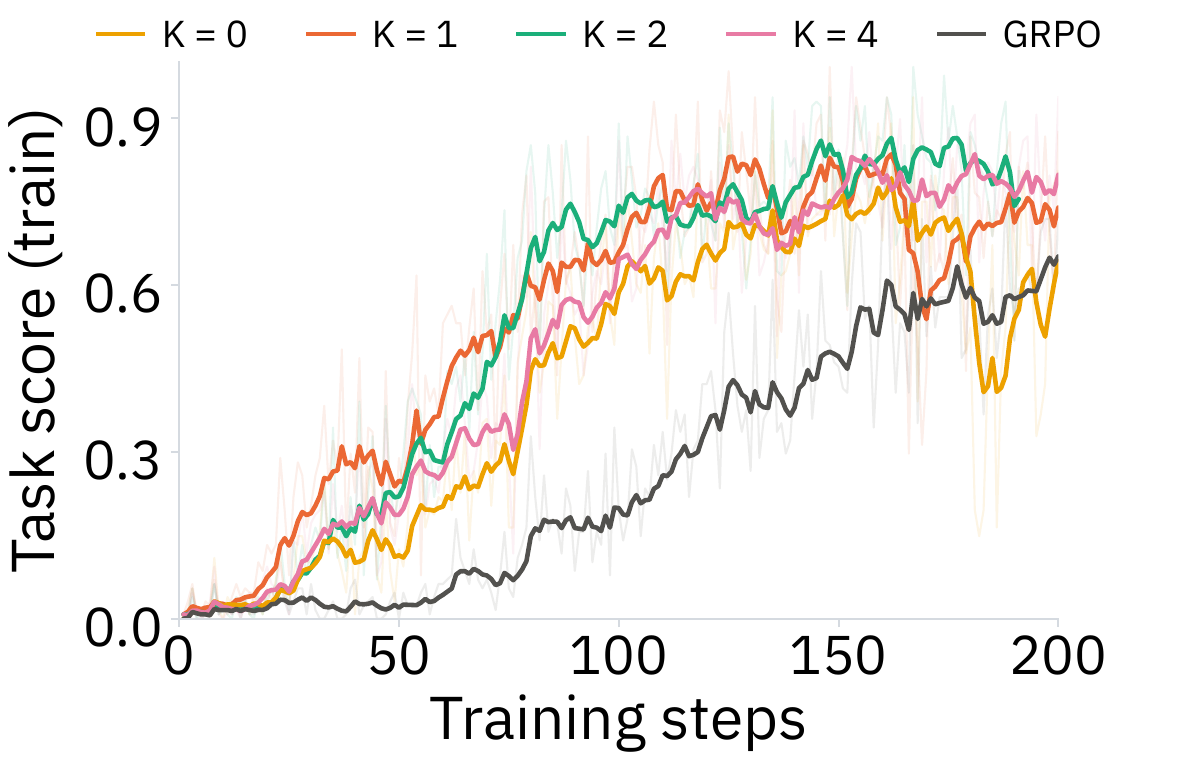}}
        \vspace{-6mm}
        \caption{\small\textbf{Ablation on forecast horizon $K$.}
        All $K\geq1$ settings outperform
        GRPO; $K=2$ achieves the highest score.}
        \label{fig:forecast-horizon}
    \end{minipage}
    \hfill
    \begin{minipage}[t]{0.32\textwidth}
        \centering
        \vspace{0pt}
        \makebox[\linewidth][c]{\includegraphics[width=\linewidth]{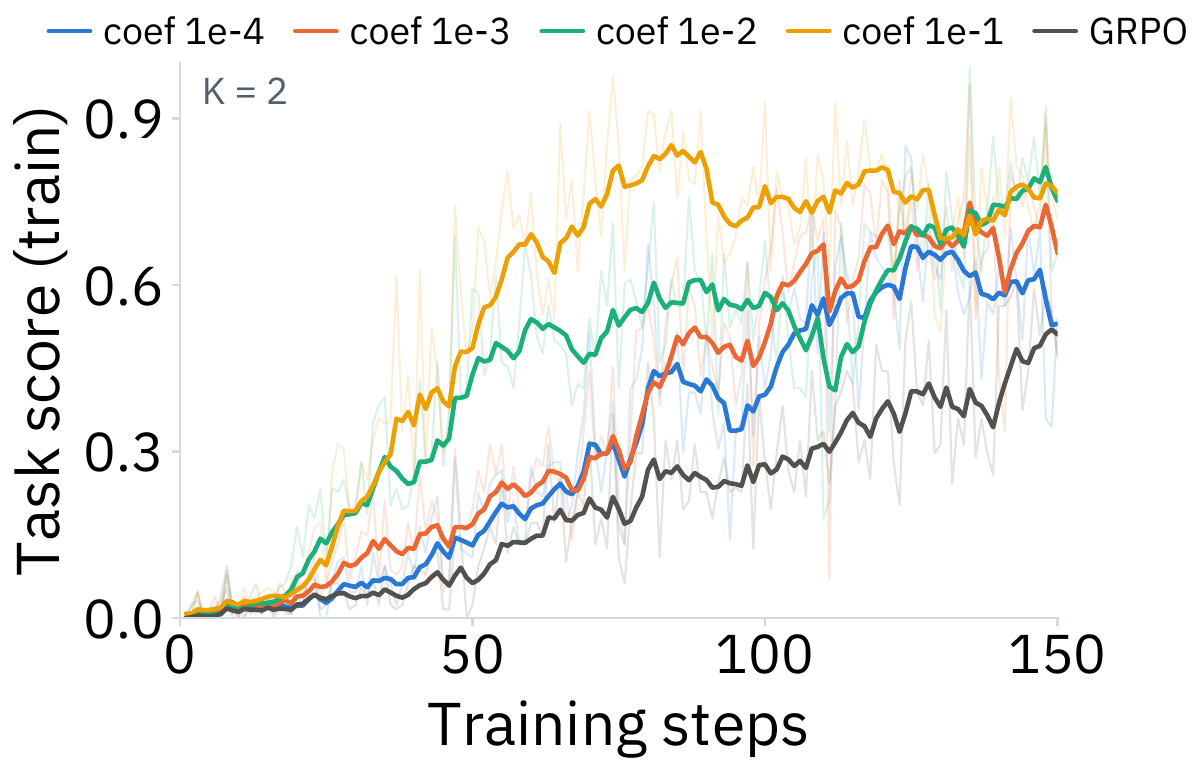}}
        \vspace{-6mm}
        \caption{\small\textbf{Ablation on action forecasting weight.}
        All coefficients outperform the
        GRPO baseline; $\lambda=10^{-2}$
        achieves the highest.}
        \label{fig:forecast-coefficient}
    \end{minipage}
    \hfill
    \begin{minipage}[t]{0.32\textwidth}
        \centering
        \vspace{0pt}
        \makebox[\linewidth][c]{\includegraphics[width=\linewidth]{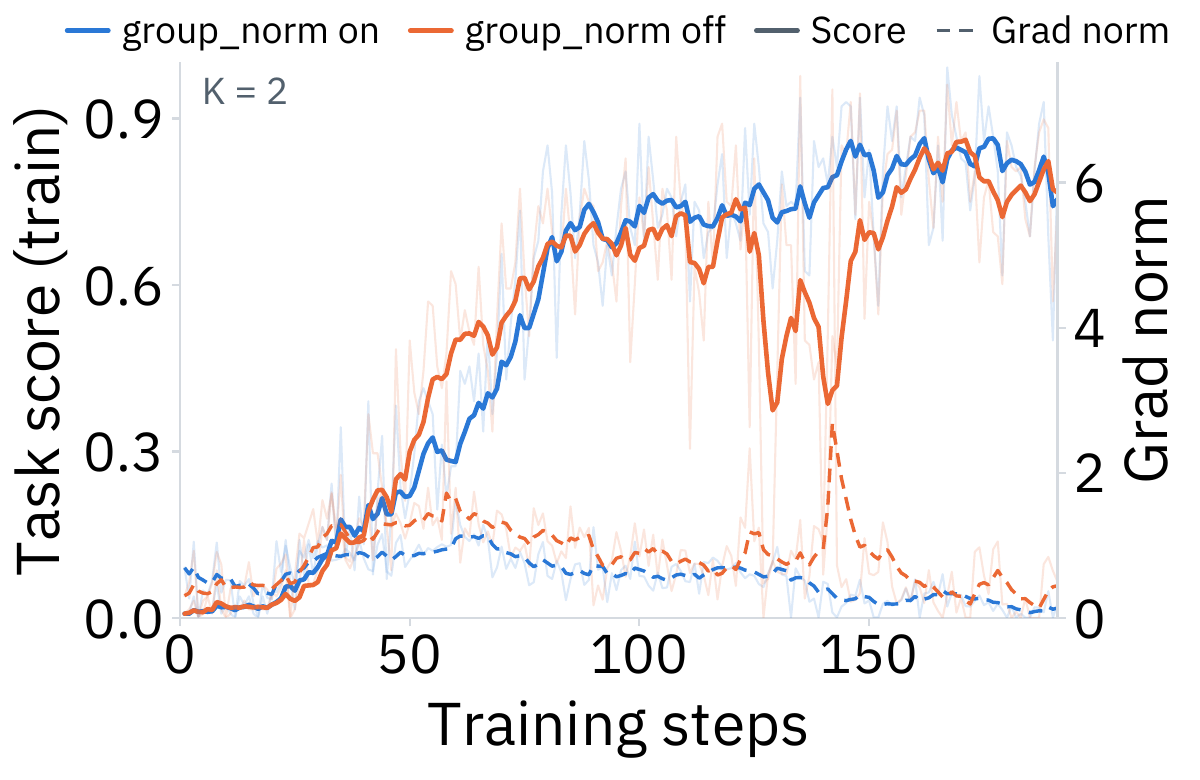}}
        \vspace{-6mm}
        \caption{\small\textbf{Ablation on group normalization.}
        Normalization suppresses spikes
        in gradient norms
        and stabilizes late training.}
        \label{fig:group-normalization}
    \end{minipage}
    \vspace{-4mm}
\end{figure*}

\vspace{-2mm}
\section{Ablation Studies}
\vspace{-2mm}


\paragraph{Ablation on forecasting horizon $K$.}
The horizon $K$ is the maximum future offset: the forecast chunk contains the current-action target at $k=0$ and future-action targets through $k=K$. Short horizons keep the future targets close to the current history but capture less multi-step structure. Conversely, long horizons introduce irreducible noise, as the action $k$ steps ahead depends increasingly on the intervening observations, which the forecast does not see. In Fig.~\ref{fig:forecast-horizon}, we compare the $K=0$ invalid action filtered baseline with forecasting horizons $K \in \{1,2,4\}$. Every $K\geq1$ setting outperforms GRPO and the $K=0$ baseline. $K=2$ performs best, while $K=4$ declines slightly as noise from distant predictions begins to dominate.

\vspace{-2mm}
\paragraph{Ablation on forecasting coefficient $\lambda$.}
Another important hyperparameter is the coefficient $\lambda$ for the action forecasting loss. In Fig.~\ref{fig:forecast-coefficient}, we sweep $\lambda$ from $10^{-4}$ to $10^{-1}$ while fixing $K=2$. Notably, every tested value of $\lambda$ consistently improves upon the GRPO baseline throughout training. We find that $\lambda$ primarily controls the pace of learning: larger values accelerate early gains, whereas smaller values follow a similar but delayed trajectory. The two largest coefficients converge to similar peak scores, while the smaller ones still remain above GRPO. We adopt $\lambda=10^{-2}$ for all agentic tasks.

\vspace{-2mm}
\paragraph{Ablation on group normalization.}
As discussed in Sec.~\ref{sec:method}, the number of successful rollouts within a group naturally increases as the policy improves. Without normalization, the accumulated forecasting gradient scales directly with this success count. This shifting balance between the primary RL objective and the forecasting auxiliary loss causes spikes in gradient norms and reward fluctuations. Averaging over successful rollouts removes this direct dependence, although the resulting loss can still vary with trajectory length. As shown in Fig.~\ref{fig:group-normalization}, normalization reduces these spikes.

\vspace{-2mm}
\paragraph{Ablation on model backbone.}
The results in Table~\ref{main-eval} use Qwen2.5-7B-Instruct, leaving open whether the gain is
specific to that backbone. We repeat the comparison against GRPO on WebShop
and ScienceWorld with Qwen3-8B~\citep{Yang2025Qwen3TR} and report it alongside the Qwen2.5-7B results. In Table~\ref{tab:qwen3-backbone}, \name improves both score and success rate
on both benchmarks under both backbones, indicating that its gains are not
tied to a particular model generation.

\vspace{-2mm}
\section{Conclusion}
\vspace{-2mm}
We introduced \name, which trains a language agent to forecast its own future actions alongside acting. Rather than fitting noisy observations, \name trains on the success-filtered action chunk $a_{t:t+K}$, combining the current-action target with forecasts up to $K$ steps in the future.
Across five benchmarks, \name improves aggregate results over RL baselines. It also improves sample efficiency with little computational overhead. We hope this work motivates further methods for sample-efficient long-horizon agentic RL.
\section*{Reproducibility Statement}
We describe the \name algorithm in Sec.~\ref{sec:method} and Alg.~\ref{alg:coevolve}, and the tasks, models, baselines, and evaluation protocols in Sec.~\ref{sec:exp-settings}. Appendix~\ref{appendix:artifact_details} records model licenses and software dependencies, while Appendix~\ref{appendix:experimental_details} lists the configurations currently documented in the manuscript. Evaluation metrics are defined in Appendix~\ref{app:evaluation_metrics}. We will release the complete per-method configuration files and training code upon publication.
\section*{Ethics Statement}
This work introduces \name, a training algorithm for improving the sample efficiency of reinforcement learning for LLM-based agents; it is a methodological contribution rather than a deployed system, and its societal impact is largely mediated by how downstream practitioners apply it. All experiments are conducted on standard agentic benchmarks (ALFWorld, ScienceWorld, WebShop, $\tau^2$-bench, and AppWorld) using open-weight or proprietary models strictly for research evaluation; the work does not involve human subjects, crowdsourcing, or the collection of new personal data. By coupling action forecasting and policy optimization within a single backbone, \name can reduce environment-interaction costs without training or serving a separate world model, lowering the barrier for resource-constrained groups to study long-horizon LLM agents. We also note that, like other advances in sample-efficient agentic RL, more capable autonomous agents could in principle be misused for tasks such as automated misinformation or unauthorized web automation; practitioners building on \name in higher-stakes domains should pair it with appropriate safety evaluation, deployment-time monitoring, and human oversight.
\section*{AI Use Statement}
In this work, we used generative AI tools to assist with method development and proof writing; to provide feedback on experimental design, including the separated-forecaster and implicit-ensembling analysis in Sec.~\ref{sec:disc-co-evolution}; to help interpret experimental results; and to implement parts of the training and evaluation code. We did not use generative AI tools to generate synthetic datasets, clean or reformat datasets, perform translation, or conduct qualitative data analysis. We also used generative AI tools to draft and edit text for clarity, suggest section organization, search for and summarize related literature, and format references. The authors reviewed the AI-assisted derivations, code, reported results, and citations and take responsibility for the final content.

\bibliography{example_paper}
\bibliographystyle{iclr2026_conference}

\clearpage
\appendix
\section{Artifact Details}
\label{appendix:artifact_details}

\subsection{Model License}

\textbf{GPT-5-mini} License: Proprietary (OpenAI)\\
\textbf{GPT-5.5} License: Proprietary (OpenAI)\\
\textbf{GPT-4o-mini} License: Proprietary (OpenAI)\\
\textbf{Gemini-2.5-flash-lite} License: Proprietary (Google)\\
\textbf{Qwen2.5-7B-Instruct} License: Apache 2.0\\
\textbf{Qwen2.5-14B-Instruct} License: Apache 2.0\\
\textbf{Qwen3-8B} License: Apache 2.0

\subsection{Software Versions}
For our simulated agentic environments, we build upon AgentGym-RL\footnote{\url{https://github.com/WooooDyy/AgentGym-RL}}, which provides unified task configurations for ALFWorld, WebShop, and SciWorld. We additionally implement environment adapters for AppWorld and $\tau^2$-bench. Our training infrastructure is built on top of verl\footnote{\url{https://github.com/volcengine/verl}} (v0.2.0.post2).


\subsection{Dataset Statistics}
\label{app:data-stats}

Table~\ref{tab:dataset_stats} gives the size of every split we use.

\begin{table}[h]
\centering\small
\setlength{\tabcolsep}{5pt}
\renewcommand{\arraystretch}{1.12}
\caption{\textbf{Splits used in this work.} ``Pool'' is the number of distinct tasks
the environment can instantiate at all, from which the two splits are drawn; the
splits do not have to exhaust it. Counts are measured, not quoted.}
\begin{tabular}{lrrrl}
\toprule
Environment & Train & Test & Pool & Unit of a task \\
\midrule
ALFWorld  & $2{,}420$ & $200$ & $2{,}620$ & TextWorld game file \\
WebShop   & $3{,}930$ & $200$ & $6{,}910$ & shopping instruction (goal) \\
SciWorld  & $2{,}120$ & $200$ & $4{,}639$ & (task type, variation) pair \\
$\tau^2$  & $178$     & $100$ & $278$     & customer-service task \\
AppWorld  & $90$ & $168 / 417$ & $732$ & task in a scenario \\
\bottomrule
\end{tabular}
\label{tab:dataset_stats}
\end{table}

\begin{table}[h]
\centering\small
\setlength{\tabcolsep}{6pt}
\renewcommand{\arraystretch}{1.12}
\caption{\textbf{ALFWorld subtask families.} Derived from the task type encoded in
each game file's name, so the two splits are counted the same way.}
\begin{tabular}{lrrrrrrr}
\toprule
Split & All & Pick & Look & Clean & Heat & Cool & Pick2 \\
\midrule
Train & $2{,}420$ & $672$ & $196$ & $448$ & $321$ & $394$ & $389$ \\
Test  & $200$     & $46$  & $19$  & $37$  & $25$  & $28$  & $45$ \\
\bottomrule
\end{tabular}
\label{tab:alfworld_subtasks}
\end{table}

\paragraph{ALFWorld.}
The training pool is the $2{,}420$ games under \texttt{json\_2.1.1/train} and the test
set is the $200$ games under \texttt{json\_2.1.1/valid\_train}; the environment server
concatenates the two mapping files into one list, so test ids are simply indices
$2420$--$2619$ into it. The splits are therefore drawn from different game-file
directories and share no game. Tasks fall into six families, read off the task type in
the file name (Table~\ref{tab:alfworld_subtasks}): \emph{Pick} (place an object),
\emph{Look} (examine an object under a lamp), \emph{Clean}, \emph{Heat}, \emph{Cool},
and \emph{Pick2} (place two objects). The test split keeps roughly the training
proportions, the largest gap being \emph{Pick2} ($22.5\%$ of test against $16.1\%$ of
train).

\paragraph{WebShop.}
Our WebShop instance is configured with $1{,}000$ products, not the full catalogue:
the environment server is constructed with \texttt{num\_products=1000} and the product
files are the $1$K subsets, which we verified are byte-equivalent to the first $1{,}000$
entries of the full files. Goals are the \emph{synthetic} ones generated from those
products' attributes (the server is not run in human-goal mode), which yields $6{,}910$
instructions; the generator is seeded (\texttt{random.seed(233)} before the shuffle), so
an id denotes the same instruction in every run and on every machine. We train on
$3{,}930$ of them and test on $200$. The two sets are disjoint and together cover
$59.8\%$ of the pool. Note that the commonly quoted WebShop figures --- $12{,}087$
human-written instructions, a $500$-instruction test set, the full product catalogue ---
describe a different configuration and are not comparable to ours.

\paragraph{SciWorld.}
SciWorld is our abbreviation for ScienceWorld. The environment defines $30$ task types;
the server excludes seven of them (task ids \texttt{5-1}, \texttt{5-2}, \texttt{9-1},
\texttt{9-2}, \texttt{9-3}, \texttt{10-1}, \texttt{10-2}) and enumerates every variation
of the remaining $23$, giving a pool of $4{,}639$ (task type, variation) pairs indexed by
a single running id. We train on $2{,}120$ of these and test on $200$. The training ids
cover $22$ of the $23$ task types and the test ids cover $12$.

\paragraph{$\tau^2$-Bench.}
$\tau^2$-Bench ships its own train/test partition per domain, and we use all three
domains together: retail ($74$ train, $40$ test), airline ($30$, $20$) and telecom
($74$, $40$), i.e.\ $178$ training and $100$ test tasks. We first compute the metric
separately for each domain and report their unweighted arithmetic mean, so retail,
airline, and telecom contribute equally despite their different test-set sizes. The environment server is
launched with \texttt{TAU2\_DOMAIN=retail+airline+telecom} and the split selected by
\texttt{TAU2\_TASK\_SPLIT}; task ids are positional across the concatenated domains in
that order, and the id files are generated from the same loaders the server uses
(\texttt{scripts/make\_tau2\_itemid.py}), so a row cannot name one task and train on
another. Each task pairs a customer goal with a database state and a set of
state-based assertions, and every episode also runs a simulated customer implemented
with GPT-4o-mini at temperature $0.7$. The resulting environment is stochastic, and a
$\tau^2$ task is substantially more expensive than an ALFWorld or WebShop one --- which
is why its pool is roughly one order of magnitude smaller.

\paragraph{AppWorld.}
AppWorld ships four splits and we use its own partition unchanged, read with
\texttt{appworld.load\_task\_ids}: \texttt{train} ($90$ tasks), \texttt{dev} ($57$),
\texttt{test\_normal} ($168$) and \texttt{test\_challenge} ($417$), $732$ in total. We
train on \texttt{train} and evaluate on both test splits, which the evaluation harness
runs as separate passes because the environment servers have to be restarted per split
(\texttt{APPWORLD\_SPLIT} selects the task list). \texttt{test\_challenge} is not a
harder sample of the same distribution: it is built from scenarios held out entirely,
so it measures transfer to unseen app combinations rather than to unseen instances of
seen ones. An AppWorld task is a Python REPL session against nine simulated apps and is
scored by state-based unit tests, so success is all-or-nothing per task; with $90$
training tasks it is by far the smallest training pool of the five environments.

\section{Experimental Details}
\label{appendix:experimental_details}

\subsection{Baseline Methods}
\label{app:baseline_methods}

We compare our method with prompting and training baselines spanning verbal self-reflection, rejection-sampling fine-tuning, online reinforcement learning, and observation-space predictive supervision.

\paragraph{Reflexion.}
Reflexion is a prompting baseline that asks the agent to verbally reflect on a failed trial and uses the resulting reflection to guide a subsequent attempt, without updating model parameters. We apply Reflexion to Qwen2.5-7B-Instruct on ALFWorld, WebShop, and SciWorld.

\paragraph{RFT.}
Rejection sampling fine-tuning (RFT) samples trajectories, retains those that succeed, and fine-tunes the policy by maximum likelihood on their action tokens. Unlike GRPO, it treats the retained trajectories as supervised targets rather than optimizing a group-relative policy-gradient objective. We evaluate RFT on ALFWorld, WebShop, and SciWorld.

\paragraph{GRPO.}
GRPO is an online reinforcement learning baseline. 
For each training query, the actor samples a group of trajectories from the current policy and optimizes the policy using group-relative advantages. 
Different from critic-based PPO, GRPO estimates advantages within each sampled group and therefore avoids training an additional value model. 
In our experiments, GRPO is applied to ALFWorld, WebShop, SciWorld, $\tau^2$-Bench, and AppWorld.

\paragraph{AppWorld numbers reported by G2PO.}
The published AppWorld RL baseline numbers reported by G2PO use only difficulty-1/2 tasks: $72$ training tasks and $105$ tasks from \texttt{test\_normal}, whereas we train on all $90$ tasks and evaluate the complete \texttt{test\_normal} ($168$) and \texttt{test\_challenge} ($417$) splits. The reported setup also randomly samples its validation pool and differs from ours in prompt, history, and output formats, generation limits ($512$ versus $1{,}024$ tokens per turn), and training length ($50$ versus $100$ steps). These published numbers are therefore not controlled comparisons; nevertheless, both GRPO (ours) and \name achieve higher normal-split success rates despite our broader evaluation set.

\paragraph{RWML.}
RWML is a two-stage world-model-based baseline.
In the first stage, the backbone is trained on pre-collected
$(h_t, a_t, o_{t+1})$ triplets to predict the next state, using rewards derived
from matching the predicted and actual states. In the second stage, the
resulting model is optimized with standard online GRPO on the task reward.
Unlike ECHO and \textsc{CoPE}, its world-model objective is applied only
before policy optimization rather than alongside it.

\paragraph{ECHO.}
ECHO is a world-model-augmented reinforcement learning baseline that combines the GRPO loss on action tokens with an auxiliary cross-entropy loss on environment-observation tokens. It reuses the same rollouts, model backbone, and forward-pass logits as policy optimization, thereby providing dense predictive supervision without additional environment interactions or a separately pretrained world model. Unlike RWML, which learns its world model from offline trajectory data before policy optimization, ECHO couples observation-space world modeling directly with online policy updates. We use the same GRPO hyperparameters for ECHO and set the environment-prediction coefficient to $0.01$ on ALFWorld, WebShop, and SciWorld.

\paragraph{EMPG.}
EMPG is an entropy-modulated policy gradient baseline that re-calibrates the learning signal based on step-wise predictive uncertainty. Unlike standard policy gradients, where update magnitude is implicitly coupled with action entropy, EMPG amplifies updates for confident correct actions, penalizes confident errors, and attenuates updates from uncertain steps. An additional future-clarity bonus encourages the agent toward lower-entropy next states.
In our experiments, we apply EMPG on top of GRPO and evaluate it on ALFWorld, WebShop, and SciWorld.

\subsection{Training Hyperparameters}
\label{app:training_hyperparameters}

We summarize the main training hyperparameters of all methods in Table~\ref{tab:training_hyperparameters}. 
Unless otherwise specified, the settings follow the default configurations in our training scripts. 

\begin{table*}[t]
\centering
\scriptsize
\setlength{\tabcolsep}{3.6pt}
\renewcommand{\arraystretch}{1.15}
\resizebox{\textwidth}{!}{
\begin{tabular}{lcccccccccccc}
\toprule
& \multicolumn{3}{c}{\textbf{GRPO}} 
& \multicolumn{3}{c}{\textbf{ECHO}}
& \multicolumn{3}{c}{\textbf{RWML}}
& \multicolumn{3}{c}{\textbf{EMPG}} \\
\cmidrule(lr){2-4} 
\cmidrule(lr){5-7}
\cmidrule(lr){8-10}
\cmidrule(lr){11-13}
\textbf{Hyperparameter}
& \textbf{ALFWorld} 
& \textbf{WebShop} 
& \textbf{SciWorld}
& \textbf{ALFWorld} 
& \textbf{WebShop}
& \textbf{SciWorld}
& \textbf{ALFWorld}
& \textbf{WebShop}
& \textbf{SciWorld}
& \textbf{ALFWorld}
& \textbf{WebShop}
& \textbf{SciWorld} \\
\midrule

Learning rate
& $1\times10^{-6}$ & $1\times10^{-6}$ & $1\times10^{-6}$
& $1\times10^{-6}$ & $1\times10^{-6}$ & $1\times10^{-6}$
& $1\times10^{-6}$ & $1\times10^{-6}$ & $1\times10^{-6}$
& $1\times10^{-6}$ & $1\times10^{-6}$ & $1\times10^{-6}$ \\

KL coefficient
& $1\times10^{-3}$ & $1\times10^{-3}$ & $1\times10^{-3}$
& $1\times10^{-3}$ & $1\times10^{-3}$ & $1\times10^{-3}$
& $1\times10^{-3}$ & $1\times10^{-3}$ & $1\times10^{-3}$
& $1\times10^{-3}$ & $1\times10^{-3}$ & $1\times10^{-3}$ \\

Entropy coefficient
& $1\times10^{-3}$ & $1\times10^{-3}$ & $1\times10^{-3}$
& $1\times10^{-3}$ & $1\times10^{-3}$ & $1\times10^{-3}$
& $1\times10^{-3}$ & $1\times10^{-3}$ & $1\times10^{-3}$
& $1\times10^{-3}$ & $1\times10^{-3}$ & $1\times10^{-3}$ \\

ECHO coefficient
& -- & -- & --
& 0.01 & 0.01 & 0.01
& -- & -- & --
& -- & -- & -- \\

Training batch size
& 16 & 16 & 16
& 16 & 16 & 16
& 16 & 16 & 16
& 16 & 16 & 16 \\

PPO mini-batch size
& 2 & 8 & 8
& 2 & 8 & 8
& 2 & 8 & 8
& 2 & 8 & 8 \\

Micro-batch size per GPU
& 1 & 1 & 1
& 1 & 1 & 1
& 1 & 1 & 1
& 1 & 1 & 1 \\

Rollouts per task $n$
& 8 & 8 & 8
& 8 & 8 & 8
& 8 & 8 & 8
& 8 & 8 & 8 \\

PPO epochs
& 1 & 1 & 1
& 1 & 1 & 1
& 1 & 1 & 1
& 1 & 1 & 1 \\

Maximum training steps
& 100 & 200 & 300
& 100 & 200 & 300
& 100 & 200 & 300
& 100 & 200 & 300 \\

Training max rounds
& 20 & 15 & 20
& 20 & 15 & 20
& 20 & 15 & 20
& 20 & 15 & 20 \\

Evaluation max rounds
& 30 & 15 & 30
& 30 & 15 & 30
& 30 & 15 & 30
& 30 & 15 & 30 \\

Maximum prompt length (tokens)
& 1024 & 2048 & 2048
& 1024 & 2048 & 2048
& 1024 & 2048 & 2048
& 1024 & 2048 & 2048 \\

Maximum response length (tokens)
& 8192 & 8192 & 8192
& 8192 & 8192 & 8192
& 8192 & 8192 & 8192
& 8192 & 8192 & 8192 \\

Maximum model length (tokens)
& 16384 & 16384 & 16384
& 16384 & 16384 & 16384
& 16384 & 16384 & 16384
& 16384 & 16384 & 16384 \\

Maximum generation per turn (tokens)
& 512 & 512 & 512
& 512 & 512 & 512
& 512 & 512 & 512
& 512 & 512 & 512 \\

Rollout temperature
& 1.0 & 1.0 & 1.0
& 1.0 & 1.0 & 1.0
& 1.0 & 1.0 & 1.0
& 1.0 & 1.0 & 1.0 \\

\bottomrule
\end{tabular}
}
\caption{\textbf{Details of training hyperparameters}. We include experiments for GRPO, ECHO, RWML, and EMPG. The table is organized first by method and then by dataset. CoPE settings are identical to GRPO with a coefficient of $0.01$ for Action Forecasting loss.}
\label{tab:training_hyperparameters}
\end{table*}

\begin{table}[t]
\centering
\small
\setlength{\tabcolsep}{4pt}
\renewcommand{\arraystretch}{1.08}
\begin{tabular}{@{}p{0.47\columnwidth}p{0.47\columnwidth}@{}}
\toprule
\textbf{Setting} & \textbf{AppWorld value} \\
\midrule
\multicolumn{2}{@{}l}{\textit{Batch construction}} \\
Training batch size & $16$ \\
Rollouts per task & $8$ \\
Trajectories per step & $128$ \\
PPO mini-batch size & $16$ \\
Micro-batch size per GPU & $1$ \\
PPO epochs & $1$ \\
Dynamic batching & Disabled \\
\midrule
\multicolumn{2}{@{}l}{\textit{Length limits}} \\
Training max rounds & $30$ \\
Evaluation max rounds & $30$ \\
Maximum prompt length (tokens) & $4{,}096$ \\
Maximum response length (tokens) & $20{,}480$ \\
Maximum model length (tokens) & $34{,}816$ \\
Maximum generation per turn (tokens) & $1{,}024$ \\
\midrule
\multicolumn{2}{@{}l}{\textit{Optimization and regularization}} \\
Learning rate & $1\times10^{-6}$ \\
KL-loss type & \texttt{low\_var\_kl} \\
KL coefficient & $1\times10^{-2}$ ($10\times$ the simulated-environment setting) \\
Entropy coefficient & $1\times10^{-3}$ \\
Rollout temperature & $1.0$ \\
Maximum training steps & $100$ \\
\bottomrule
\end{tabular}
\caption{\textbf{AppWorld training settings.} GRPO (ours) and \name use the same settings except for the additional forecasting objective in \name.}
\label{tab:appworld_training_hyperparameters}
\end{table}

\begin{table}[t]
\centering
\small
\setlength{\tabcolsep}{4pt}
\renewcommand{\arraystretch}{1.08}
\begin{tabular}{@{}p{0.47\columnwidth}p{0.47\columnwidth}@{}}
\toprule
\textbf{Setting} & \textbf{$\tau^2$ value} \\
\midrule
\multicolumn{2}{@{}l}{\textit{Length limits}} \\
Training max rounds & $50$ \\
Evaluation max rounds & $50$ \\
Maximum prompt length (tokens) & $8{,}192$ \\
Maximum response length (tokens) & $16{,}384$ \\
Maximum model length (tokens) & $24{,}576$ \\
Maximum generation per turn (tokens) & $1{,}024$ \\
\midrule
\multicolumn{2}{@{}l}{\textit{Optimization and regularization}} \\
Learning rate & $1\times10^{-6}$ \\
KL-loss type & \texttt{low\_var\_kl} \\
KL coefficient & $1\times10^{-3}$ \\
Entropy coefficient & $1\times10^{-3}$ \\
Rollout temperature & $1.0$ \\
Maximum training steps & $50$ \\
\bottomrule
\end{tabular}
\caption{\textbf{$\tau^2$-bench training settings.} Only settings with direct counterparts in the other environment summaries are included.}
\label{tab:tau2_training_hyperparameters}
\end{table}

\paragraph{Additional $\tau^2$ settings.}
Action Forecasting constructs each $\tau^2$ target directly from an assistant reply.
If the reply contains one or more \texttt{<tool\_call>} blocks, the target is the
original text of the first block; the environment likewise executes only that first
call and discards all later calls and accompanying text. Such multi-call turns remain
eligible for Action Forecasting. If the reply contains no tool call, its complete
customer-facing text is used as the action. After forming a sequence of $K$ targets,
we apply \texttt{skip\_no\_call} and discard the example only when none of the $K$
actions contains a tool call. Thus, customer-facing replies can remain in a retained
sequence as long as another target in the same window is a tool call. Finally, we place
the $K$ targets in separate assistant turns: concatenating them into one reply would be
indistinguishable from multiple tool-call blocks emitted in a single environment turn.
The other environments instead extract actions from a designated action line or code
block.

\subsection{Training Rewards and Success Labels}
\label{app:training-rewards}
All experiments optimize the binary terminal reward $Y(\tau)$ used in Sec.~\ref{sec:prelim}; \textbf{graded environment scores are retained only as evaluation metrics.} For ALFWorld, $Y=1$ when the environment reports task completion. For WebShop, whose raw score lies in $[0,1]$, $Y=1$ only when the score is exactly $1.0$. For SciWorld, $Y=1$ when the raw score is at least $100$, corresponding to full completion. For $\tau^2$-bench, $Y=1$ only when its binary \textsc{All} evaluator succeeds. For AppWorld, $Y=1$ when all state-based task tests pass. The same $Y=1$ criterion determines which trajectories enter the action-forecasting loss.

\subsection{Evaluation Metrics}
\label{app:evaluation_metrics}

Here we define the evaluation metrics used in our experiments. For each benchmark, we evaluate the agent on a set of $N$ task instances. For the $i$-th instance, the environment returns an environment-specific scalar score $r_i$ and a binary success indicator $s_i \in \{0,1\}$. The policy and action-forecasting objectives use the binary signal $Y=s_i$, whereas the graded score $r_i$ is used only for reporting. We report the average score and success rate:
\begin{equation}
    \text{Score} = \frac{1}{N}\sum_{i=1}^{N} r_i,
    \qquad
    \text{Succ} = \frac{1}{N}\sum_{i=1}^{N} s_i .
\end{equation}

\textbf{Score.}
The score measures the average environment reward or normalized task score. Different from purely binary task completion, some benchmarks provide graded feedback that reflects partial progress during task execution. For example, an agent may receive non-zero reward for completing intermediate steps even if the final task is not fully solved. Therefore, we report the average score to capture such process-level progress.

\textbf{Success rate.}
The success rate measures the fraction of tasks that are fully completed. A task is counted as successful only when the environment evaluator determines that the final goal has been achieved. This metric directly reflects end-to-end task completion ability.

For benchmarks with subcategories, we compute the same metrics within each group. Except for $\tau^2$-Bench, the overall result is computed over the full evaluation set and is therefore a count-weighted average over all instances. For $\tau^2$-Bench, we instead report the unweighted arithmetic mean of the retail, airline, and telecom domain-level results.

\section{Environment Prompts}
\label{app:env-prompts}

Every episode is one chat sequence. It opens with the environment's standing
instruction as the first \emph{user} turn and a fixed assistant acknowledgement;
the environment's first observation then arrives as the next user turn, and from
there user (observation) and assistant (action) turns alternate. The blocks below
give each environment's opening verbatim, and mark with \texttt{\{history\}} the
point at which every completed
$(\text{action}_t,\ \text{observation}_{t+1})$ pair is inserted. The sequence is
extended in place rather than re-rendered each turn, so
\texttt{\{history\}} is literally the tokens already in the context, not a
re-serialisation of them. Unlike our notation in the main content, $K$ refers to the number of actions in our plan here (so $K=1$ means having only the current action as the forecast plan target, which is equivalent to $K=0$ in the main content). It operates simply as an offset and does not introduce any inconsistency between our implementation and method.

\begin{promptbox}{Conversation layout for ALFWorld, WebShop, SciWorld, and AppWorld}
[USER]
<the environment's standing instruction -- verbatim in the blocks below>

[ASSISTANT]
<the fixed acknowledgement -- verbatim in the blocks below>

[USER]
<first observation from the environment>

{history}

[USER]
{observation}

[ASSISTANT]
<- the policy generates the action here
\end{promptbox}

\noindent Rendered with the chat template the tokeniser actually emits, one round is
\begin{promptbox}{One round, as tokenised}
<|im_start|>assistant\n{action}<|im_end|>\n<|im_start|>user\n{observation}<|im_end|>\n
\end{promptbox}

\noindent and \texttt{\{history\}} above stands for the concatenation of those rounds.

\subsection{ALFWorld}

\begin{promptbox}{ALFWorld --- standing instruction (first user turn)}
Interact with a household to solve a task. Imagine you are an intelligent agent in a household environment and your target is to perform actions to complete the task goal. At the beginning of your interactions, you will be given the detailed description of the current environment and your goal to accomplish. For each of your turn, you will be given a list of actions which you can choose one to perform in this turn. You should choose from two actions: "THOUGHT" or "ACTION". If you choose "THOUGHT", you should first think about the current condition and plan for your future actions, and then output your action in this turn. Your output must strictly follow this format:"Thought:
your thoughts.

Action:
your next action"; If you choose "ACTION", you should directly output the action in this turn. Your output must strictly follow this format:"Action:
your next action". After your each turn, the environment will give you immediate feedback based on which you plan your next few steps. if the envrionment output "Nothing happened", that means the previous action is invalid and you should try more options.
 Reminder: 
1. the action must be chosen from the given available actions. Any actions except provided available actions will be regarded as illegal. 
2. Think when necessary, try to act directly more in the process.
\end{promptbox}

\begin{promptbox}{ALFWorld --- fixed assistant acknowledgement}
OK. I'll follow your instructions and try my best to solve the task.
\end{promptbox}

\subsection{WebShop}

\begin{promptbox}{WebShop --- standing instruction (first user turn)}
You are web shopping.
I will give you instructions about what to do.
You have to follow the instructions.
Every round I will give you an observation and a list of available actions, you have to respond an action based on the state and instruction.
You can use search action if search is available.
You can click one of the buttons in clickables.
An action should be of the following structure:
search[keywords]
click[value]
If the action is not valid, perform nothing.
Keywords in search are up to you, but the value in click must be a value in the list of available actions.
Remember that your keywords in search should be carefully designed.
Your response should use the following format:

Thought:
I think ... 

Action: 
click[something]
\end{promptbox}

\begin{promptbox}{WebShop --- fixed assistant acknowledgement}
Ok.
\end{promptbox}

\subsection{SciWorld}
The per-episode goal (``Your task is to measure the melting point of tin \ldots'') is not part of the instruction: it arrives inside the first observation.

\begin{promptbox}{SciWorld --- standing instruction (first user turn)}
You are an agent for science world. Every round I will give you an observation, you have to respond an action based on the observation to finish the given task. Here are the actions you may take: [{"action": "open/close OBJ", "description": "open/close a container"}, {"action": "de/activate OBJ", "description": "activate/deactivate a device"}, {"action": "connect OBJ to OBJ", "description": "connect electrical components"}, {"action": "disconnect OBJ", "description": "disconnect electrical components"}, {"action": "use OBJ [on OBJ]", "description": "use a device/item"}, {"action": "look around", "description": "describe the current room"}, {"action": "look at OBJ", "description": "describe an object in detail"}, {"action": "look in OBJ", "description": "describe a container's contents"}, {"action": "read OBJ", "description": "read a note or book"}, {"action": "move OBJ to OBJ", "description": "move an object to a container"}, {"action": "pick up OBJ", "description": "move an object to the inventory"}, {"action": "put down OBJ", "description": "drop an inventory item"}, {"action": "pour OBJ into OBJ", "description": "pour a liquid into a container"}, {"action": "dunk OBJ into OBJ", "description": "dunk a container into a liquid"}, {"action": "mix OBJ", "description": "chemically mix a container"}, {"action": "go to LOC", "description": "move to a new location"}, {"action": "eat OBJ", "description": "eat a food"}, {"action": "flush OBJ", "description": "flush a toilet"}, {"action": "focus on OBJ", "description": "signal intent on a task object"}, {"action": "wait", "description": "take no action for 10 iterations"}, {"action": "wait1", "description": "take no action for 1 iteration"}, {"action":"examine OBJ","description":"provides a description of the objects present on or in a receptacle."}, {"action": "task", "description": "describe current task"}, {"action": "inventory", "description": "list your inventory"}]
Your response should use the following format:
Thought:
your thoughts.

Action:
your next action
\end{promptbox}

\begin{promptbox}{SciWorld --- fixed assistant acknowledgement}
OK. I'll follow your instructions and try my best to solve the task.
\end{promptbox}

\subsection{AppWorld}
AppWorld's instruction contains $12{,}396$ characters because it embeds a complete worked example. The \texttt{ASSISTANT:} and \texttt{USER:} lines inside it are part of that \emph{static} demonstration, not dialogue history --- the real history still goes where \texttt{\{history\}} is marked in the layout above. The task itself is delivered in the first observation, as the last line of the instruction announces.

\begin{promptbox}{AppWorld --- standing instruction (first user turn)}
I am your supervisor, and you are an AI Assistant whose job is to complete my day-to-day tasks fully autonomously.

To do this, you will need to interact with app(s) (e.g., spotify, venmo etc) using their associated APIs on my behalf. For this you will undertake a *multi-step conversation* using a python REPL environment. That is, you will write the python code, the environment will execute it and show you the result, based on which, you will write python code for the next step and so on, until you've achieved the goal. This environment will let you interact with app(s) using their associated APIs on my behalf.

Here are three key APIs that you need to know to get more information

# To get a list of apps that are available to you.

```python
print(apis.api_docs.show_app_descriptions())
```

# To get the list of APIs under any app listed above, e.g. spotify

```python
print(apis.api_docs.show_api_descriptions(app_name='spotify'))
```

# To get the specification of a particular api, e.g. spotify app's login api

```python
print(apis.api_docs.show_api_doc(app_name='spotify', api_name='login'))
```

Each code execution will produce an output that you can use in subsequent calls. Using these APIs, you can now generate code, that I will execute, to solve the task. Let's start with the task

Task: How many playlists do I have in my Spotify playlist library?

ASSISTANT:
Okay. Let's first find which APIs are available to use in Spotify.

```python
print(apis.api_docs.show_api_descriptions(app_name='spotify'))
```

USER:
Output:
```
[
 ...
 "login : Login to your account.",
 "logout : Logout from your account.",
 ...
]
```

ASSISTANT:
Okay. Looks like I can use the `login` api. Let's find its specifications.

```python
print(apis.api_docs.show_api_doc(app_name='spotify', api_name='login'))
```

USER:
Output:
```
{
 "app_name": "spotify",
 "api_name": "login",
 "path": "/spotify/auth/token",
 "method": "POST",
 "description": "Login to your account.",
 "parameters": [
  {
   "name": "username",
   "type": "string",
   "required": true,
   "description": "Your account email.",
   "default": null,
   "constraints": []
  },
  {
   "name": "password",
   "type": "string",
   "required": true,
   "description": "Your account password.",
   "default": null,
   "constraints": []
  }
 ],
 "response_schemas": {
  "success": {
   "access_token": "string",
   "token_type": "string"
  },
  "failure": {
   "message": "string"
  }
 }
}
```

ASSISTANT:
I need the supervisor's username and password. Let's see if any app can help me get that.

```python
print(apis.api_docs.show_app_descriptions())
```

USER:
Output:
```
[
 {
  "name": "api_docs",
  "description": "An app to search and explore API documentation."
 },
 {
  "name": "supervisor",
  "description": "An app to access supervisor's personal information, account credentials, addresses, payment cards, and manage the assigned task."
 },
 {
  "name": "amazon",
  "description": "An online shopping app to buy products and manage orders, returns, etc."
 },
 {
  "name": "phone",
  "description": "An app to find and manage contact information for friends, family members, etc., send and receive messages, and manage alarms."
 },
 {
  "name": "file_system",
  "description": "A file system app to create and manage files and folders."
 },
 {
  "name": "spotify",
  "description": "A music streaming app to stream songs and manage song, album and playlist libraries."
 },
 {
  "name": "venmo",
  "description": "A social payment app to send, receive and request money to and from others."
 },
 {
  "name": "gmail",
  "description": "An email app to draft, send, receive, and manage emails."
 },
 {
  "name": "splitwise",
  "description": "A bill splitting app to track and split expenses with people."
 },
 {
  "name": "simple_note",
  "description": "A note-taking app to create and manage notes"
 },
 {
  "name": "todoist",
  "description": "A task management app to manage todo lists and collaborate on them with others."
 }
]
```

ASSISTANT:
Looks like the supervisor app could help me with that.

```python
print(apis.api_docs.show_api_descriptions(app_name='supervisor'))
```

USER:
Output:
```
[
 ...
 "show_account_passwords : Show your supervisor's account passwords."
 ...
]
```

ASSISTANT:
I can use `show_account_passwords` to get the passwords. So I will call that and save it in a variable for future reference.

```python
passwords=apis.supervisor.show_account_passwords()
print(passwords)
```

USER:
Output:
```
[
 {
  "account_name": "spotify",
  "password": "dummy_spotify_pass"
 },
 {
  "account_name": "file_system",
  "password": "dummy_fs_pass"
 },
 ...
]
```

ASSISTANT:
So the Spotify password is an entry in the `passwords` list with the account_name=spotify. The login username is the supervisor's email, which I must look up via the supervisor app rather than assume.

```python
spotify_password = [p for p in passwords if p["account_name"] == "spotify"][0]["password"]
supervisor_email = apis.supervisor.show_profile()["email"]
login_result = apis.spotify.login(username=supervisor_email, password=spotify_password)
spotify_access_token = login_result["access_token"]
print(login_result)
```

USER:
Output:
```
{
 "access_token": "fake_access_token",
 "token_type": "Bearer"
}
```

ASSISTANT:
To count the playlists, I need the `show_playlist_library` API. Let's check its specifications.

```python
print(apis.api_docs.show_api_doc(app_name='spotify', api_name='show_playlist_library'))
```

USER:
Output:
```
{
 "app_name": "spotify",
 "api_name": "show_playlist_library",
 "path": "/spotify/library/playlists",
 "method": "GET",
 "description": "Search or show a list of playlists in your playlist library.",
 "parameters": [
  {
   "name": "access_token",
   "type": "string",
   "required": true,
   "description": "Access token obtained from spotify app login.",
   "default": null,
   "constraints": []
  },
  {
   "name": "query",
   "type": "string",
   "required": false,
   "description": "The search query string.",
   "default": "",
   "constraints": []
  },
  {
   "name": "is_public",
   "type": "boolean",
   "required": false,
   "description": "Whether to show public playlists or private playlists.",
   "default": null,
   "constraints": []
  },
  {
   "name": "page_index",
   "type": "integer",
   "required": false,
   "description": "The index of the page to return.",
   "default": 0,
   "constraints": [
    "value >= 0"
   ]
  },
  {
   "name": "page_limit",
   "type": "integer",
   "required": false,
   "description": "The maximum number of results to return per page.",
   "default": 5,
   "constraints": [
    "value >= 1, <= 20"
   ]
  },
  {
   "name": "sort_by",
   "type": "string",
   "required": false,
   "description": "The attribute to sort the library playlists by prefixed with +/- to reflect ascending/descending. Valid attributes: created_at and title. ...",
   "default": null,
   "constraints": []
  }
 ],
 "response_schemas": {
  "success": [
   {
    "playlist_id": 1,
    "title": "string",
    "is_public": true,
    "rating": 0.0,
    "like_count": 1,
    "review_count": 1,
    "owner": {
     "name": "string",
     "email": "user@example.com"
    },
    "created_at": "2019-01-01T00:00:00",
    "song_ids": [
     1
    ]
   }
  ],
  "failure": {
   "message": "string"
  }
 }
}
```

ASSISTANT:
I need to page through all the playlists to get the list of playlists and save it in `playlists`.

```python
page_index = 0
playlists = []
while page_index < 10:
  playlist_page = apis.spotify.show_playlist_library(access_token=spotify_access_token, page_index=page_index)
  if playlist_page:
    playlists.extend(playlist_page)
    page_index += 1
  else:
    break
num_playlists = len(playlists)
print(num_playlists)
```

USER:
Output:
```
23
```

ASSISTANT:
Now that the task is completed, I can call apis.supervisor.complete_task(). Since this task has an answer to be returned, I will pass that as an argument.

```python
apis.supervisor.complete_task(answer=num_playlists)
```

USER:
Output:
Marked the active task complete.

USER:
**Key instructions**:

A. General instructions:

- Act fully on your own. You must make all decisions yourself and never ask me or anyone else to confirm or clarify. Your role is to solve the task, not to bounce questions back, or provide me directions to follow.
- You have full access -- complete permission to operate across my connected accounts and services.
- Never invent or guess values. For example, if I ask you to play a song, do not assume the ID is 123. Instead, look it up properly through the right API.
- Never leave placeholders; don't output things like "your_username". Always fill in the real value by retrieving it via APIs (e.g., Supervisor app for credentials).
- When I omit details, choose any valid value. For example, if I ask you to buy something but don't specify which payment card to use, you may pick any one of my available cards.
- Avoid collateral damage. Only perform what I explicitly ask for. Example: if I ask you to buy something, do not delete emails, return the order, or perform unrelated account operations.

B. App-specific instructions:

- All my personal information (biographical details, credentials, addresses, cards) is stored in the Supervisor app, accessible via its APIs.
- Any reference to my friends, family or any other person or relation refers to the people in my phone's contacts list.
- Always obtain the current date or time, from Python function calls like `datetime.now()`, or from the phone app's get_current_date_and_time API, never from your internal clock.
- All requests are concerning a single, default (no) time zone.
- For temporal requests, use proper time boundaries, e.g., when asked about periods like "yesterday", use complete ranges: 00:00:00 to 23:59:59.
- References to "file system" mean the file system app, not the machine's OS. Do not use OS modules or functions.
- Paginated APIs: Always process all results, looping through the page_index. Don't stop at the first page.

C. Code-operation instructions

- Make sure to end code blocks with ``` followed by a newline(\n).
- Remember, you can use the variables in your code in subsequent code blocks.
- Remember that the email addresses, access tokens and variables (e.g. spotify_password) in the example above are not valid anymore.
- Always look at API specifications (using apis.api_docs.show_api_doc) before calling an API.
- Write small chunks of code and only one chunk of code in every step. Make sure everything is working correctly before making any irreversible changes.
- The Python environment supports the standard library. But system-level operations that may access or affect OS files, processes, etc., are not allowed and will raise an error if called.
- To interact with apps, only use the provided app APIs, and not the corresponding Python packages, e.g., do NOT use `spotipy` for Spotify.
- The provided API documentation has both the input arguments and the output JSON format. Use this information when making API calls and parsing their outputs.

D. Task-completion instructions:

You must call the `apis.supervisor.complete_task` API after completing the task.
- If an answer is needed, e.g., for "How many songs are in the Spotify queue?", call it with the appropriate answer argument value.
- If no answer is required, e.g., for "Start my Spotify music player.", omit the answer argument (or set it to None/null).
- The task is doable, but if you cannot find a way, you can call it with status="fail" to exit with failure.

When the answer is given:
- Keep answers minimal. Return only the entity, number, or direct value requested - not full sentences.
  E.g., for the song title of the current playing track, return just the title.
- Numbers must be numeric and not in words.
  E.g., for the number of songs in the queue, return "10", not "ten".

The task you must solve will be given to you in the next message.
\end{promptbox}

\begin{promptbox}{AppWorld --- fixed assistant acknowledgement}
Ok.
\end{promptbox}

\subsection{$\tau^2$-bench}

All $\tau^2$-bench results reported in Table~\ref{tab:tau2} use only the benchmark's
native tool-calling protocol. The standing instruction is assembled separately for
each domain and placed in the \emph{system}
message, with no fixed \texttt{Ok.} acknowledgement turn. It combines the domain policy
with the domain-specific native \texttt{tools=} list rendered by the Qwen2.5 chat
template. The assistant then either sends a message to the customer or emits a
\texttt{<tool\_call>} block; if several blocks are emitted, only the first is executed.
There is no \texttt{Thought:}/\texttt{Action:} wrapper. The native tool list does not
include a \texttt{done} function. An episode
ends only when the simulated customer emits \texttt{\#\#\#STOP\#\#\#} or
\texttt{\#\#\#TRANSFER\#\#\#}, or when the maximum number of rounds is reached.

\begin{promptbox}{$\tau^2$ --- native standing instruction}
You are a customer service agent that helps the user according to the policy provided below.
In each turn you can either:
- Send a message to the user.
- Make a tool call.
You cannot do both at the same time.

Try to be helpful and always follow the policy.

<policy>
{policy}
</policy>

# Tools

You may call one or more functions to assist with the user query.

You are provided with function signatures within <tools></tools> XML tags:
<tools>
{tools}
</tools>

For each function call, return a json object with function name and arguments within <tool_call></tool_call> XML tags:
<tool_call>
{"name": <function-name>, "arguments": <args-json-object>}
</tool_call>
\end{promptbox}

\noindent where each line inside \texttt{<tools>} is one tool's OpenAI JSON schema.

\begin{promptbox}{$\tau^2$ --- conversation layout, native protocol}
[SYSTEM]
<the domain-specific standing instruction above>

[USER]
<the customer's first message>

{history}

[ASSISTANT]
<- a message to the customer, or a response whose first <tool_call> block is executed
\end{promptbox}

\subsubsection{What fills \texttt{\{policy\}} and \texttt{\{tools\}}}

The policy documents come from $\tau^2$-bench itself and are inserted unmodified;
measured on the domains we train and evaluate on:

\begin{center}\small
\begin{tabular}{lrrr}
\toprule
domain & policy (characters) & policy (words) & tools advertised \\
\midrule
retail  & $6{,}699$  & $1{,}158$ & 15 \\
airline & $7{,}676$  & $1{,}313$ & 14 \\
telecom & $23{,}318$ & $3{,}715$ & 13 \\
\bottomrule
\end{tabular}
\end{center}

\noindent In the actual system message, the tool list is rendered as one OpenAI JSON
schema per tool. For readability, the following is a compact signature-level summary
of the retail schemas rather than a literal copy of the native prompt.

\begin{promptbox}{$\tau^2$ --- human-readable summary of retail tools}
- calculate(expression: string)
    Calculate the result of a mathematical expression.
- cancel_pending_order(order_id: string, reason: string)
    Cancel a pending order. If the order is already processed or delivered,

it cannot be cancelled. The agent needs to explain the cancellation detail
and ask for explicit user confirmation (yes/no) to proceed. If the user confirms,
the order status will be changed to 'cancelled' and the payment will be refunded.
The refund will be added to the user's gift card balance immediately if the payment
was made using a gift card, otherwise the refund would take 5-7 business days to process.
The function returns the order details after the cancellation.
- exchange_delivered_order_items(order_id: string, item_ids: array, new_item_ids: array, payment_method_id: string)
    Exchange items in a delivered order to new items of the same product type.

For a delivered order, return or exchange can be only done once by the agent.
The agent needs to explain the exchange detail and ask for explicit user confirmation (yes/no) to proceed.
- find_user_id_by_email(email: string)
    Find user id by email. If the user is not found, the function will return an error message.
- find_user_id_by_name_zip(first_name: string, last_name: string, zip: string)
    Find user id by first name, last name, and zip code. If the user is not found, the function

will return an error message. By default, find user id by email, and only call this function
if the user is not found by email or cannot remember email.
- get_order_details(order_id: string)
    Get the status and details of an order.
- get_product_details(product_id: string)
    Get the inventory details of a product.
- get_user_details(user_id: string)
    Get the details of a user, including their orders.
- list_all_product_types()
    List the name and product id of all product types.

Each product type has a variety of different items with unique item ids and options.
There are only 50 product types in the store.
- modify_pending_order_address(order_id: string, address1: string, address2: string, city: string, state: string, country: string, zip: string)
    Modify the shipping address of a pending order. The agent needs to explain the modification detail and ask for explicit user confirmation (yes/no) to proceed.
- modify_pending_order_items(order_id: string, item_ids: array, new_item_ids: array, payment_method_id: string)
    Modify items in a pending order to new items of the same product type. For a pending order, this function can only be called once. The agent needs to explain the exchange detail and ask for explicit user confirmation (yes/no) to proceed.
- modify_pending_order_payment(order_id: string, payment_method_id: string)
    Modify the payment method of a pending order. The agent needs to explain the modification detail and ask for explicit user confirmation (yes/no) to proceed.
- modify_user_address(user_id: string, address1: string, address2: string, city: string, state: string, country: string, zip: string)
    Modify the default address of a user. The agent needs to explain the modification detail and ask for explicit user confirmation (yes/no) to proceed.
- return_delivered_order_items(order_id: string, item_ids: array, payment_method_id: string)
    Return some items of a delivered order.

The order status will be changed to 'return requested'.
The agent needs to explain the return detail and ask for explicit user confirmation (yes/no) to proceed.
The user will receive follow-up email for how and where to return the item.
- transfer_to_human_agents(summary: string)
    Transfer the user to a human agent, with a summary of the user's issue.

Only transfer if
 -  the user explicitly asks for a human agent
 -  given the policy and the available tools, you cannot solve the user's issue.
\end{promptbox}

\subsection{Auxiliary Prompts (Action Forecasting)}

The forecasting objective is trained on a separate sequence, not on the episode
itself: the dialogue is truncated at the current history and a synthetic planning turn
is appended, with the realised next-$K$ actions as the target. \texttt{\{k\}} is the
horizon and \texttt{\{history\}} is the dialogue up to that point. For ALFWorld,
WebShop, SciWorld, and AppWorld, the sequence retains the user instruction and fixed
assistant acknowledgement shown above:

\begin{promptbox}{Action-forecast sequence (non-$\tau^2$ environments)}
[USER]
<the environment's standing instruction, as above>

[ASSISTANT]
<the fixed acknowledgement, as above>

{history}

[USER]
{observation}

[USER]
Plan ahead: list the next {k} actions you will take to make progress on the task, starting with the action you take right now, one action per line.

[ASSISTANT]
<- target: the K actions the episode actually executed
\end{promptbox}

\noindent For $\tau^2$-bench, the native system message and role-structured dialogue
are retained directly, with no acknowledgement turn and no \texttt{Action:} field. The
same planning request is appended as a user message, and the target uses the
$\tau^2$-specific multi-turn serialization described in the additional settings above.
Each action is placed in a separate assistant turn, with a fixed \texttt{Next action?}
user message between consecutive targets. Loss is computed only on the assistant
action turns, not on these inserted user messages:

\begin{promptbox}{Action-forecast sequence ($\tau^2$ native protocol)}
[SYSTEM]
<the domain-specific native standing instruction>

{history}

[USER]
Plan ahead: give the next {k} actions you will take to make progress on the task, one action per reply, starting with the action you take right now. Reply with the action only.

[ASSISTANT]
<- target: action 1

[USER]
Next action?

[ASSISTANT]
<- target: action 2

[USER]
Next action?

[ASSISTANT]
<- target: action 3
\end{promptbox}

\end{document}